\documentclass[letterpaper]{article} 
\usepackage{aaai2027}  
\usepackage[hyphens]{url}  
\usepackage{graphicx} 
\usepackage{amsmath}   
\usepackage{amssymb}   
\usepackage{natbib}  
\usepackage{caption} 
\usepackage{algorithm}
\usepackage{algorithmic}

\usepackage{newfloat}
\usepackage{listings}
\DeclareCaptionStyle{ruled}{labelfont=normalfont,labelsep=colon,strut=off} 
\floatstyle{ruled}
\newfloat{listing}{tb}{lst}{}
\floatname{listing}{Listing}

\usepackage{booktabs}

\usepackage{multirow}
\usepackage{url}

\nocopyright 

\title{The Geometry of Flow-Matching Uncertainty: A Cost-free Uncertainty Proxy \\ and Application in Flow-based VLA Failure Detection}
\author {
    Ziyang Rao\textsuperscript{\rm 1,4},
    Yiren Zhao\textsuperscript{\rm 1,4},
    Ben Fei\textsuperscript{\rm 3},
    Yandong Guo\textsuperscript{\rm 4},
    Weiyu Guo\textsuperscript{\rm 3,4}\corresponding,
    Hui Xiong\textsuperscript{\rm 1,2}\corresponding
}
\affiliations {
    \textsuperscript{\rm 1}HKUST(GZ) \quad
    \textsuperscript{\rm 2}HKUST \quad
    \textsuperscript{\rm 3}CUHK \quad
    \textsuperscript{\rm 4}AI² Robotics
}

\begin{document}

\maketitle

\begin{abstract}
Flow matching (FM) has become a popular action head paradigm for modern embodied models. 
However, as a conditional generative model, it does not explicitly expose its inherent \emph{uncertainty}, producing faulty action chunks even when it misinterprets the scene or encounters out-of-distribution (OOD) inputs. 
Therefore, determining when an FM-generated action can be trusted is essential for safe deployment, yet existing uncertainty estimation methods on real-time control suffer from several issues: extra training budget, high computational overhead, and low generalization ability.
In this work, we provide a geometric interpretation of FM uncertainty in the velocity field, showing that uncertainty manifests as deviation from an ideal affine-isotropic contraction field.
Building on this observation, we introduce \emph{denoising acceleration} ($\mathrm{accel}$), a highly-generalizable and cost-free uncertainty proxy that measures the bending of the denoising trajectory from a \emph{single} forward pass, without additional model evaluations, training, or resampling.
We theoretically and empirically demonstrate that $\mathrm{accel}$ is a faithful proxy for FM uncertainty and further test its utility in online failure detection.
Results show that $\mathrm{accel}$ identifies failing rollouts well before termination, matching or even outperforming costly resampling- and training-based baselines across settings under realistic deployment budget. Furthermore, $\mathrm{accel}$ localizes \emph{where} the uncertainty physically lies on a real bimanual robot at no extra cost. Code and demos available at: \url{https://github.com/rrrrrrzy/fm-geometry}.

\end{abstract}


\section{Introduction}
\label{sec:1_intro}

\begin{flushright}
\begin{minipage}{\linewidth}
\small\itshape
``There is a goal, but no way; what we call the way is hesitation.''

\raggedleft
--- Franz Kafka, \textit{The Zürau Aphorisms}
\end{minipage}
\end{flushright}

\vspace{0.5em}

Flow matching (FM) \citep{lipman2023flow} has recently become a popular paradigm for the action heads of model embodied policies like vision language action models (VLA) or world action model (WA). 
It generates precise action trajectories, capture complex multimodal action distributions
\citep{DBLP:journals/corr/abs-2410-24164, braun2024riemann},
and enable lower-latency sampling than diffusion policies
\citep{DBLP:conf/rss/ChiFDXCBS23}.
Besides powerful capabilities, FM heads are compatible with a wide range of model architectures, e.g.\ mixture of experts structure of $\pi$ \citep{DBLP:journals/corr/abs-2410-24164, DBLP:journals/corr/abs-2504-16054}, dual-system architecture of GR00T \citep{DBLP:journals/corr/abs-2503-14734}, or coupled with world model components \citep{DBLP:journals/corr/abs-2504-02792,zhou2026tau0wm}.

\begin{figure}[!t]
    \centering
    \includegraphics[width=\columnwidth]{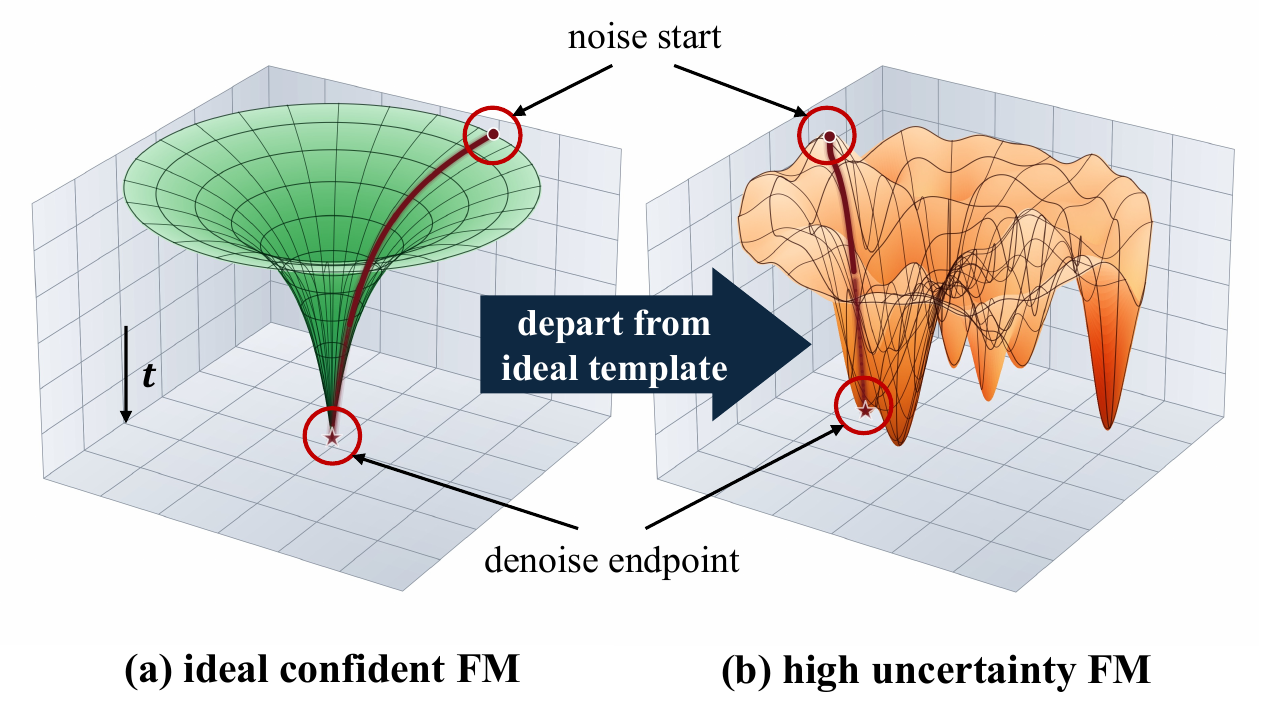}
    \caption{%
    \textbf{The geometry of FM velocity field, schematic.}
    Schematic geometry of \textbf{(a)} an ideally confident and a \textbf{(b)} highly uncertain FM. The fields are visualized as following the downhill directions of an illustrative landscape.
    Dark red curves are denoising trajectories
    from independent noise.}
    \label{fig:teaser}
\end{figure}

However, like many generative models, FM policies are susceptible to a hallucination-like failure mode: they can fail \emph{silently}, producing smooth and seemingly executable action chunks even when they misunderstand the task, encounter out-of-distribution (OOD) inputs, or predict physically infeasible outcomes \citep{agia2024sentinel, xu2025faildetect}.
Unlike classical controllers equipped with explicit constraints \citep{khatib1986real} and collision checks \citep{schulman2014motion}, the black-box nature of FM models does not provide a trust-worthy signal of whether its generated actions is correct.

Existing studies of uncertainty in flow-based VLAs either train additional parameters to estimate variance from the model state \citep{han2026uaflow}, or comprise \emph{Monte Carlo-style} approaches that resample outputs and quantify their spread or consistency \citep{agia2024sentinel, roemer2025fiper}.
However, these methods are either non-generalizable or computationally expensive, particularly consequential for real-time manipulation.
Recent work like FreeHunch \citep{DBLP:conf/iclr/RissanenHS25} and the closed-form covariance in \citep{DBLP:journals/corr/abs-2605-00941} reveal a new direction and imply the connection between FM field geometry and uncertainty, 
and yet require tens of forward-equivalent computations \citep{DBLP:journals/corr/abs-2605-00941}.
This limitation motivate a central question: \emph{how to estimate FM model uncertainty with minimal overhead?}

To address this question, we first investigate the nature of certainty in FM and ground it in the geometry of the flow field.
We then introduce \emph{denoising acceleration}, denoted by $\mathrm{accel}$, as a faithful, highly-generalizable and cost-free proxy for FM uncertainty and apply it to online failure detection.
In summary, this paper makes the following contributions:
\begin{itemize}
    \item \textbf{A geometric definition of FM uncertainty.}
    We characterize the velocity field of a high certainty FM as approximately radial and define FM uncertainty as its deviation from this ideal template, and visualize different uncertainty geometry with a toy and a real FM model.

    \item \textbf{A cost-free online proxy for FM uncertainty.}
    We propose \emph{denoising acceleration} ($\mathrm{accel}$) as a proxy for FM uncertainty. It reads off the bend from a single denoising pass without additional training, ensembling, or resampling.
    Through theoretical and empirical analyses, we demonstrate its agreement with computationally expensive resampling-based uncertainty reference.

    \item \textbf{A free lunch for closed-loop failure detection.}
    We use $\mathrm{accel}$ as a zero-cost, real-time failure detector for FM-based VLAs that requires no extra training or labelled data.
    Under a realistic deployment budget, it achieves performance comparable or even better than state-of-the-art detectors with additional training or resampling, suggesting $\mathrm{accel}$'s utility in capturing rich uncertainty signals.

    \item \textbf{Localizing uncertainty in the action space.}
    More case studies show $\mathrm{accel}$ tells not only \emph{whether} a chunk is uncertain but \emph{which} part on a real robot attributing errors to specific action coordinates, also at zero extra cost.
\end{itemize}
\section{The Geometric Nature of FM Uncertainty}
\label{sec:2_geometry}

\subsection{Preliminaries}  

\begin{figure*}[!t]
    \centering
    \includegraphics[width=\textwidth]{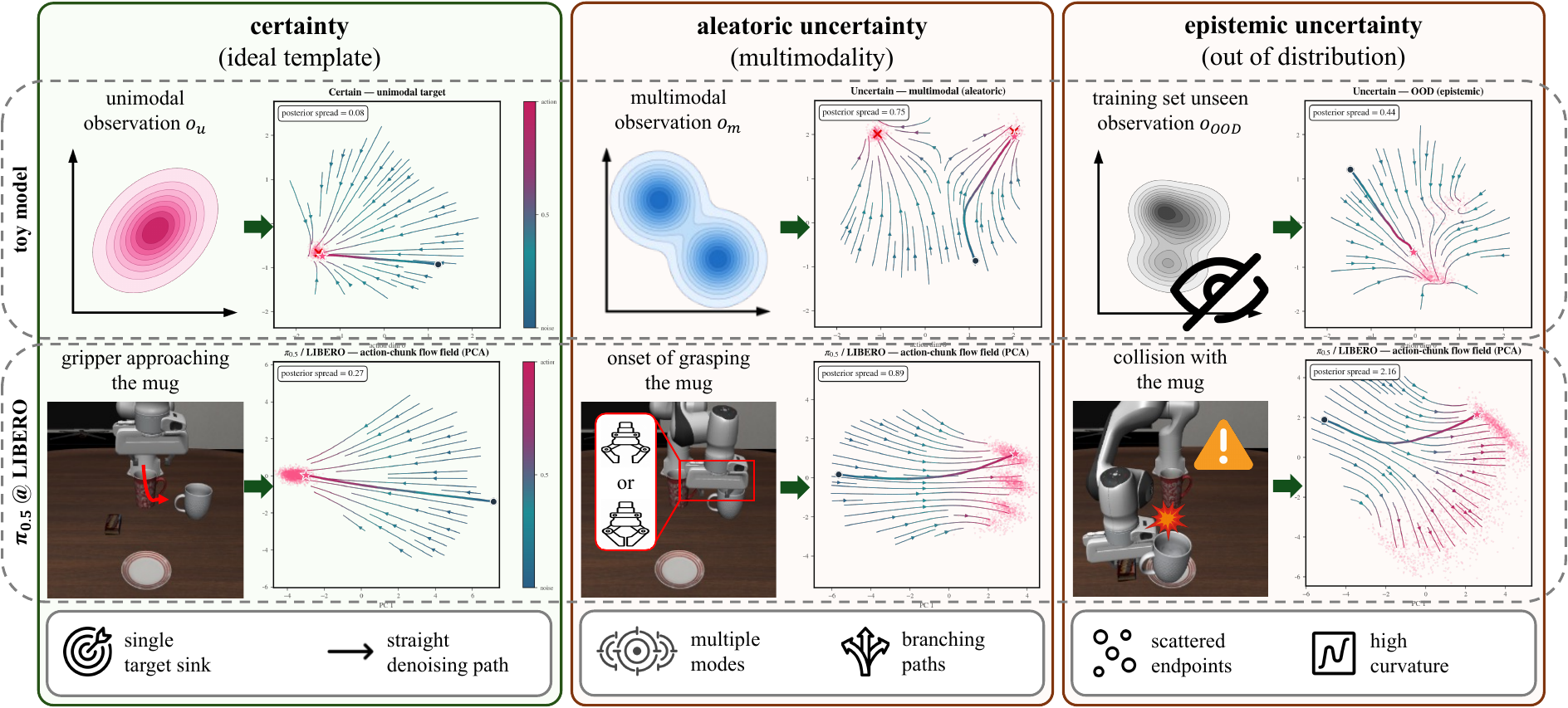}
    \caption{\textbf{Field geometry of FM uncertainty in toy model (top) and $\pi_{0.5}$ policy (bottom).}
    Each column stands for: \emph{certainty} (unimodal), \emph{aleatoric} uncertainty (multimodality) and \emph{epistemic} uncertainty (OOD). Streamlines trace the velocity field. A highlighted denoising trajectory $x_0\!\to\!x_1$ runs from noise start $\bullet$ to clean-action target $\star$. The resampled endpoints are labeled in light pink.
    \textbf{Left:} a certain, unimodal field is an affine-isotropic sink with near-straight trajectories into a tight endpoint cloud.
    \textbf{Center:} an aleatoric field displays significant endpoint clustering and trajectory bends.
    \textbf{Right:} an epistemic, OOD condition generates chaotic field sharply curved trajectories and widely scattered endpoints.}
    \label{fig:toy_flowfield}
\end{figure*}

\paragraph{Setup.}
Consider a conditional flow matching (CFM) model with independent Gaussian noise. 
With the flattened action chunk as a state $x\in\mathbb{R}^{D}$ and a forward schedule $s\in[0,1]$, the interpolant is
\begin{equation}
x_s = s\,x_1 + (1-s)\,x_0,\qquad x_0\sim\mathcal{N}(0,I),
\label{eq:interp}
\end{equation}
where $s{=}0$ is pure noise $x_0$ and $s{=}1$ is a clean action $x_1\sim p(x_1\mid o)$ drawn from the conditional target distribution given observation $o$, independently of $x_0$.
Conditioned on $o$, the network is trained toward the \emph{marginal velocity}, i.e.\ the conditional-expectation field of the per-sample velocity $x_1-x_0$,
\begin{equation}
v(x,s) \;=\; \mathbb{E}\!\left[x_1-x_0 \,\middle|\, x_s=x\right].
\label{eq:vdef}
\end{equation}
Eliminating the noise via $x_0=(x_s-s\,x_1)/(1-s)$ gives $x_1-x_0=(x_1-x_s)/(1-s)$ and thus
\begin{equation}
v(x,s) \;=\; \frac{\mathbb{E}[x_1\mid x_s=x]-x}{1-s},
\label{eq:vmean}
\end{equation}
which represents the FM velocity as a scaled vector \emph{from the current state toward the posterior mean} $\mathbb{E}[x_1\mid x_s=x]$. This step uses only the interpolant and independence; Gaussianity enters just once, in \eqref{eq:tweedie-mean}.

\paragraph{Defining certainty.}
Similar to the idea of entropy, we define FM certainty by how concentrated the posterior is. 
Denote that the posterior target distribution $p(x_1\mid o)$ has mean $a^{\star}=\mathbb{E}[x_1\mid o]$ and covariance $\Sigma=\operatorname{Cov}(x_1\mid o)$. 
The condition $o$ is considered \emph{maximally certain} when the posterior collapses to a point mass, $\Sigma=0$, and increasingly \emph{uncertain} as its dispersion grows. 
In other word, given fixed condition $o$, an ideal certain FM head generates exactly the identical action $a^{\star}$ despite the input noise. 

\subsection{What Does a Certain FM Field Look Like?}

We begin with the ideal certain FM field by taking the maximal limit $\Sigma\to0$, the endpoint posterior collapses to a point mass $\mathbb{P}(x_1=a^\star\mid o)=1$. 
Hence $\mathbb{E}[x_1\mid x_s=x]=a^{\star}$ for \emph{every} $x$ and $s$. Substituting into \eqref{eq:vmean} gives
\begin{equation}
v(x,s) \;=\; \frac{a^{\star}-x}{1-s} \;=\; \frac{1}{s-1}\,(x-a^{\star}),
J:=\frac{\partial v}{\partial x}=\frac{I}{s-1}.
\label{eq:affine}
\end{equation}
We call a field \emph{radial} about a center $c$ if $v(x)=-g(x)\,(x-c)$ with a scalar gain $g(x)>0$, i.e.\ every field vector points along the ray toward $c$. 
Equation~\eqref{eq:affine} shows that an ideal certain FM field is a special radial field with an isotropic gain $g=1/(1-s)$, an \emph{affine isotropic contraction} whose Jacobian is a negative multiple of the identity matrix. 
Integrating the denoising ODE $x’=v(x,s)$ from $x(0)=x_0$ gives the closed form solution
\begin{equation}
x(s)=a^{\star}+(1-s)\,(x_0-a^{\star}),
\label{eq:straightline}
\end{equation}
a straight segment from $x_0$ to $a^{\star}$ traversed with \emph{constant} velocity $v=a^{\star}-x_0$. And hence we have:

\begin{quote}
\noindent\textbf{Theorem 1. (Qualitative)}
\emph{If the target posterior is maximally certain ($\Sigma=0$), then the FM field is an affine-isotropic sink toward $a^\star$ with $J=-I/(1-s)$, and every denoising trajectory is a straight line into $a^{\star}$ traversed at constant speed, with zero acceleration. \\(high certainty $\Rightarrow$ affine-isotropic field)}
\label{theo:1}
\end{quote}

This is the geometric anchor of our paper, as illustrated in Figure \ref{fig:teaser}, it establishes the FM geometry induced by maximal certainty.
Meanwhile its contrapositive suggests that any departure from this affine-isotropic template rules out maximal certainty, thereby motivating an exciting question: \emph{can we measure FM uncertainty by measuring the extent to which a field departs from this ideal geometry?}

\subsection{How Off-Template Geometry Reveals Uncertainty?}

For any posterior distribution, differentiating \eqref{eq:vmean} and applying the second-order Tweedie identity for Gaussian channel
\[
x_s\mid x_1\sim \mathcal{N}(s x_1,(1-s)^2I)
\]
yields
\begin{equation}
\frac{\partial}{\partial x}\mathbb{E}[x_1\mid x_s=x]
=
\frac{s}{(1-s)^2}
\operatorname{Cov}(x_1\mid x_s=x).
\label{eq:tweedie-mean}
\end{equation}
Combining \eqref{eq:tweedie-mean} with \eqref{eq:vmean}, we obtain an important equation
\begin{equation}
\boxed{
\operatorname{Cov}(x_1\mid x_s=x)
=
\frac{(1-s)^3}{s}
\left(
J+\frac{I}{1-s}
\right),s\in(0,1).
}
\label{eq:tweedie-jacobian}
\end{equation}

The bracketed term $J+\tfrac{I}{1-s}$ is exactly the excess of the field Jacobian over the affine-isotropic template $-I/(1-s)$ of \eqref{eq:affine}, a measure of how far the field departs from that template, it is exactly proportional to the conditional endpoint covariance, i.e.\ the uncertainty measure.

It also gives the \emph{converse} of Theorem~1, the template is attained \emph{only} at $\Sigma=0$, so a nonzero excess implies $\Sigma\neq0$.
We now enhance the qualitative Theorem~1 to a quantitative version by comparing $v$ with the ideal affine field
\[
v_{\mathrm{aff}}(x,s)=\frac{a^\star-x}{1-s}.
\]
From \eqref{eq:vmean},
\[
v(x,s)-v_{\mathrm{aff}}(x,s)
=
\frac{\mathbb{E}[x_1\mid x_s=x]-a^\star}{1-s}.
\]
Taking expectation over $x_s$ and using the law of total covariance,
\begin{equation}
\mathbb{E}_{x_s}
\left[
\left\lVert
v-v_{\mathrm{aff}}
\right\rVert^2
\right]
=
\frac{
\operatorname{tr}
\operatorname{Cov}
\left(
\mathbb{E}[x_1\mid x_s]
\right)
}{(1-s)^2}
\le
\frac{\operatorname{tr}\Sigma}{(1-s)^2}.
\label{eq:global-bound}
\end{equation}
Equivalently (in fact an exact decomposition, Appendix~\ref{app:field-complements}),
\begin{equation}
\operatorname{tr}\Sigma
\ge
(1-s)^2
\mathbb{E}_{x_s}
\left[
\left\lVert
v-v_{\mathrm{aff}}
\right\rVert^2
\right].
\label{eq:uncertainty-lower-bound}
\end{equation}

\begin{quote}
\noindent\textbf{Theorem 2. (Quantitative)}
\emph{For $s\in(0,1)$, any local deviation
$J+I/(1-s)\neq0$ at even a single state implies nonzero target uncertainty $\Sigma\neq0$. Moreover the mean-square field deviation from the affine template lower-bounds the posterior covariance. \\(off-template $\Rightarrow$ non-zero uncertainty)}
\end{quote}

\subsection{How FM Uncertainty Appears in Practice?}
\label{sec:2_toy}
To visualize the geometry across uncertainty regimes, we compare a controllable toy model with a real FM policy. Both exhibit consistent geometric patterns of FM uncertainty despite their different scales.

\paragraph{Setup.}
For the toy model, we train a small conditional flow matching network $v_\theta(x,s,o)$ over 2-D action distributions $A(o)$ conditioned on an observation $o$. 
The training set contains two types of observations: unimodal observations $o_u$ paired with simple Gaussian action targets, and multimodal observations $o_m$ paired with Gaussian-mixture targets, such that a single observation is compatible with multiple desired actions. 
This setup captures how an FM policy maps observations like images, language instructions, and robot states to action distributions with varying degrees of ambiguity.
For the real policy, we finetune $\pi_{0.5}$ \citep{DBLP:journals/corr/abs-2504-16054}, a state-of-the-art FM model on LIBERO \citep{DBLP:conf/nips/LiuZGFLZS23}.

Figure~\ref{fig:toy_flowfield} shows the conditioned FM velocity fields of both the toy model and $\pi_{0.5}$ and
the fields of $\pi_{0.5}$ are projected onto their first two principal components.
The visualization reveals the geometry of a certain FM field and distinguishes two sources of uncertainty, i.e.\ aleatoric and epistemic.

\paragraph{Certainty.}
The left column provides a textbook illustration of the certain geometry described by Theorem~1.
The toy model receives an observation $o$ with a unimodal target, 
while $\pi_{0.5}$ receives a state in which the arm is approaching the mug to grasp it.
In both cases, the next action is unambiguous, and the models produce a nearly affine field, 
where the streamlines converge to a single sink like Figure \ref{fig:teaser}(a), the sampled trajectories run almost \emph{straight} from noise to action, and their endpoints form a tight cluster.

\paragraph{Aleatoric uncertainty.} 
The middle panel illustrates aleatoric uncertainty which refers to irreducible ambiguity in the data where an observation is genuinely consistent with multiple actions, i.e.\ a multimodal target distribution.
For $\pi_{0.5}$, multimodality mostly appears near \emph{physical contact}, e.g.\ at the onset of grasping the mug, when the model must decide \emph{if and when} to close the gripper.
The corresponding FM fields split into distinct basins associated with different action modes, and the trajectories \emph{bend} as they commit to one branch rather than proceeding straight toward a target.

\paragraph{Epistemic uncertainty.}
The right panel showcases epistemic uncertainty which stems from a lack of knowledge about the out-of-distribution (OOD) condition. 
The inputs $o$ are unseen in training stage, for $\pi_{0.5}$, such conditions often occur in failure cases involving unexpected collisions or jitter. 
The OOD FM fields turns chaotic and generate \emph{sharply curved} trajectories.
They depart from the ideal template not because the true posterior is multimodal
but because the models must extrapolate beyond their training support.

These visual patterns further confirms the geometric nature Theorems~1-2 predict. 
Notably, the trajectory curvature read off a single forward integration seems to carry rich information about field uncertainty. 
In the next section, we turn this observation into a practical proxy for FM uncertainty.

\section{Acceleration as an Uncertainty Proxy}
\label{sec:3_accel}

Section~\ref{sec:2_geometry} identified the geometric signature of FM uncertainty, i.e.\ the deviation from the ideal template encodes the posterior covariance. 
While this characterization makes the uncertainty operationally measurable, 
extracting field-level quantities requires many field evaluations or posterior resampling, 
incurring precisely the overhead we seek to avoid.

This section closes the loop by estimating uncertainty from a single denoising trajectory, 
already available from the standard forward pass. 
We call this cost-free proxy \emph{denoising acceleration} ($\mathrm{accel}$) 
and establish its faithfulness with theoretical and empirical evidence.

\subsection{A Trajectory-level Acceleration-based Proxy}

The ideal certain field produces a straight denoising path with constant velocity and zero acceleration. 
\textbf{Any bend in the denoising path is a visible symptom of the field being off-template}.
Therefore, we build our proxy on the trajectory level geometric property. 
We call it the \emph{denoising acceleration} $\mathrm{accel}$:
the total variation of velocity along a \emph{prefix} $[0,\tau]$ of the denoise path
(i.e.\ the accumulated magnitude of velocity change), normalized by the mean speed on that prefix so that the read is scale-free,

\begin{equation}
\mathrm{accel}_\tau
=
\frac
{\displaystyle\int_0^\tau
\left\lVert
v'(s)
\right\rVert
\,ds}
{
\displaystyle\frac{1}{\tau}\int_0^\tau
\left\lVert
v(s)
\right\rVert
\,ds
},
\qquad \tau\in(0,1),
\label{eq:accel-cont}
\end{equation}

whose discrete estimator over the first $p$ of the $T$ Euler steps is

\begin{equation}
\boxed{
\mathrm{accel}_p = \frac{p\sum_{t=1}^{p-1}\bigl\lVert
v_{t}-v_{t-1}\bigr\rVert}{\sum_{t=0}^{p-1}{\lVert v_t\rVert}}},
\label{eq:accel-disc}
\end{equation}

where velocity $v_t$ is the FM denoising output at every step $t=0,1,\dots,T-1$ and generate a trajectory $x_0, x_1,\dots,x_{T}$ from noise to target action with Euler step $x_{t+1} = x_{t} + v_t/T$. The prefix length $p\in\{2,\dots,T\}$ (continuously $\tau=p/T$) sets how many leading steps the read integrates over; the full-path proxy reads the entire trajectory, $\mathrm{accel}:=\mathrm{accel}_T$.
Note that we say the path ``bends'', but $v'=x''$ is actually a velocity variation, not an intrinsic curvature.

Given a forward denoising pass, the discrete $\mathrm{accel}_p$ and its full-path $\mathrm{accel}$ can be easily retrieved with zero extra cost, as summarized in Algorithm~\ref{alg:accel}.

\begin{algorithm}[tb]
\caption{Denoising Acceleration $\mathrm{accel}_p$}
\label{alg:accel}
\textbf{Input}: observation $o$, FM model $v_\theta$, denoise steps $n{=}T$, prefix length $p\le n$ (default $p=n$)\\
\textbf{Output}: scale-free proxy $\mathrm{accel}_p$
\begin{algorithmic}[1] 
\STATE $J \gets 0$, \; $S \gets 0$, \; $\mathrm{accel} \gets 0$
\STATE sample noise $x_0 \sim \mathcal{N}(0, I)$
\FOR{$t=0$ to $n-1$}
\STATE $v_t \gets v_\theta(x_t, o, s_t)$, \quad $s_t \gets t/n$ \hfill \# FM forward pass
\STATE $x_{t+1} \gets x_t + v_t/n$ \hfill \# Euler step
\IF{$t \le p-1$}
\STATE $S \gets S + \lVert v_t\rVert$
\IF{$t \ge 1$}
\STATE $J \gets J + \lVert v_t - v_{t-1}\rVert$
\STATE $\mathrm{accel} \gets J \,/\, \bigl(S/(t+1)\bigr)$ \hfill \# running $\mathrm{accel}_{t+1}$
\ENDIF
\ENDIF
\ENDFOR
\STATE \textbf{return} $\mathrm{accel}_p = \mathrm{accel}$
\end{algorithmic}
\end{algorithm}

\subsection{Theoretical Faithfulness to FM Uncertainty}

We now justify that this free geometric read is a \emph{faithful} proxy for the field-level uncertainty characterized in Section~\ref{sec:2_geometry}. We present the main results and proof sketches here, with full proofs deferred to the Appendix.

\paragraph{The connection to posterior covariance.}
The proxy measures how much the denoising path bends, so we first compute the path acceleration by differentiating \eqref{eq:vmean},
\begin{equation}
v'(s)=\frac{m_{\mathrm{path}}'(s)}{1-s},\qquad m_{\mathrm{path}}(s):=\mathbb{E}[x_1\mid x_s=x(s)],
\label{eq:traj-identity-body}
\end{equation}
where $m_{\mathrm{path}}$ is the posterior mean the denoiser is currently aiming at. 
This implies that motion toward a \emph{fixed} target contributes no acceleration, so the path bends \emph{if and only if} the posterior mean it is chasing is itself moving. We further demonstrate its connection to the endpoint covariance read along the path, i.e.\ Section~\ref{sec:2_geometry}'s off-template excess,
\begin{equation}
v'(s)=\frac{\partial_s m}{1-s}\;+\;\frac{s}{(1-s)^3}\,\operatorname{Cov}\bigl(x_1\mid x_s=x(s)\bigr)\,v(s).
\label{eq:accel-cov}
\end{equation}
Where Section~\ref{sec:2_geometry} measured the departure of the whole \emph{field} from the template, $\mathrm{accel}$ integrates it along one trajectory, reading the covariance through its action on $v$ --- how a single pass proxies a field-level quantity, and the limit of what it sees.

\paragraph{Two critical properties of a faithful proxy.} 
As a faithful proxy, $\mathrm{accel}$ should satisfy two key properties: zero baseline and (local) monotonicity.

\noindent\emph{Zero baseline.}
For a maximally certain FM field, i.e.\ one with zero uncertainty, $\mathrm{accel}$ should be $0$.
This follows directly from Theorem~1: when $\Sigma=0$ the velocity is constant, so $\mathrm{accel}_p=0$ for every prefix.
Euler reproduces $v_{t+1}=v_t$ identically, so this is \emph{exact} at any $T$ (Prop.~A.2).

\noindent\emph{Local monotonicity.}
Within an operational regime, consider two FM fields with $\Sigma_a=\sigma_a^2\Sigma_0$ and $\Sigma_b=\sigma_b^2\Sigma_0$ sharing a covariance shape, if $\sigma_a > \sigma_b$ and both are small in the sense made precise below,
then $\mathbb{E}\bigl[\mathrm{accel}^{(a)}\bigr] > \mathbb{E}\bigl[\mathrm{accel}^{(b)}\bigr]$.
Writing the endpoint covariance as $\Sigma=\sigma^2\Sigma_0$ with a fixed shape $\Sigma_0$, we expand \eqref{eq:traj-identity-body} on a prefix that $1-\tau\gg\sigma$ and get
\begin{equation}
\mathbb{E}_{x_0}\bigl[\mathrm{accel}_\tau\bigr]=\kappa(\tau)\,\sigma^2+O(\sigma^3),
\label{eq:leading-order-body}
\end{equation}
where
\begin{equation}
\kappa(\tau)=\tfrac{(1-\tau)^{-2}-1}{2}\,
\mathbb{E}_{x_0}\!\left[\tfrac{\lVert\Sigma_0 x_0\rVert}{\lVert x_0-a^\star\rVert}\right].
\end{equation}
$\mathbb{E}_{x_0}\bigl[\mathrm{accel}_\tau\bigr]$ is strictly positive whenever $\Sigma_0\neq0$.
Thus, the expected proxy departs from zero strictly monotonically and, to leading order, linearly in the posterior variance.
Therefore greater uncertainty yields a larger expected $\mathrm{accel}$, providing the rank faithfulness required for detection.
The slope rises with $\tau$ while the window $\sigma\ll1-\tau$ shrinks, which predicts the \emph{interior} optimal prefix of Table~\ref{tab:accel_rho} (Appendix~\ref{app:prefix-accel-rho}).

\noindent\textbf{Note:} theoretical faithfulness of $\mathrm{accel}$ is local and holds in expectation and under constraints.
Thus, we further empirically support the utility of $\mathrm{accel}$ as an uncertainty proxy in the following section.

\subsection{Empirical Evidence Across Diverse Settings}
We further empirically confirm that $\mathrm{accel}$ is highly correlated to FM uncertainty, i.e.\ the target divergence $\Sigma$ (throughout, ``divergence'' means endpoint \emph{spread}, not $\nabla\!\cdot\!v$).

\begin{figure}[!t]
    \centering
    \includegraphics[width=\linewidth]{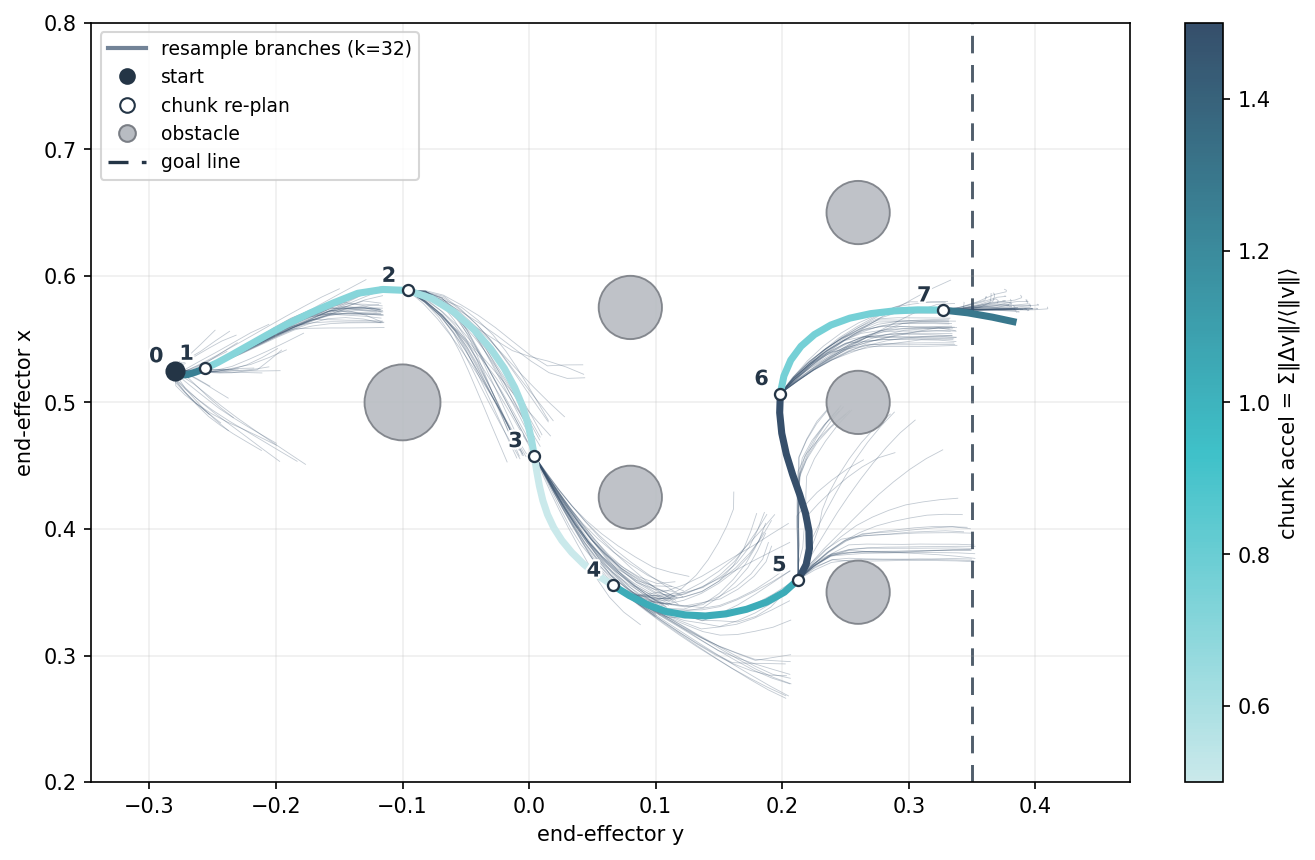}
    \caption{\textbf{Visualization of $\mathrm{accel}$ score and resampled chunks on D3IL.}
    The thick curve shows the closed-loop end-effector trajectory, colored by the chunk-level $\mathrm{accel}$.
    At each point, thin gray curves show resampled action chunks from the same observation.}
    \label{fig:accel_rho}
\end{figure}

\paragraph{Setup.}
We define the uncertainty measure as the posterior divergence among Monte-Carlo resamples. 
At every closed-loop generation step we fix the observation $o$, resample $K$ candidate action chunks from the FM head with different noise seeds, and take $D_{\mathrm{resample}}$ to be the \emph{largest} pairwise $L_2$ distance among the $K$ standardized candidates.
This is a pushforward diameter rather than $\operatorname{tr}\Sigma$, but scales as $\sigma$ and so is rank-equivalent to it at fixed $K$ and shape (Appendix~\ref{app:faithfulness}).
We report the Spearman rank correlation $\rho$ between the divergence reference and $\mathrm{accel}$ across various models and benchmarks.

\noindent\emph{Models.} We evaluate 4 flow matching VLA models across different architectures.
$\mathbf{\pi_{0.5}}$ \citep{DBLP:journals/corr/abs-2504-16054} employs a MoE style structure to connect the VLM backbone to a lightweight FM action expert. 
\textbf{GR00T N1.7} \citep{DBLP:journals/corr/abs-2503-14734} adopts a dual-system action cascade architecture to decouple high-level reasoning and real time action generation.
\textbf{VLA-JEPA} \citep{sun2026vlajepa} is augmented with a V-JEPA2 \citep{DBLP:journals/corr/abs-2506-09985} based latent world model that captures action-relevant state transitions. 
\textbf{SmolVLA} \citep{DBLP:journals/corr/abs-2506-01844} is a compact size model (0.24B) with comparable performance. 
And our \textbf{toy model} from \ref{sec:2_toy}.

\noindent\emph{Benchmarks.} The models are finetuned and evaluated on 3 benchmarks varying in task complexity and action spaces. 
\textbf{D3IL} \citep{DBLP:conf/iclr/JiaBJRDLN24} is a simple and controlled testbed for multimodal behavior learning. We focus on its \emph{Avoiding} task, where a robot must avoid six obstacles and reach a finish line.
\textbf{LIBERO} \citep{DBLP:conf/nips/LiuZGFLZS23, fei2025liberoplus} comprises 4 different tabletop manipulation suites and test understanding and manipulation capabilities.
\textbf{Robocasa} \citep{DBLP:conf/rss/NasirianyMZPLJM24, nasiriany2026robocasa365} features challenging and diverse kitchen environments. We use \emph{Atomic-Seen} split containing 18 single-skill manipulation tasks evaluated in held-out target kitchens.

\begin{table}[!t]
\centering
\caption{\textbf{Pooled chunk-level Spearman $\rho$ between $\mathrm{accel}$ and resampled divergence of $n$ total chunks.} $\rho_{\mathrm{full}}$ reports the correlation of the whole denoise path $\mathrm{accel}_T$ and $\rho_{\mathrm{best}}$ the highest $\rho$ of $\mathrm{accel}_{p^{\star}}$ from the first $p^{\star}$ of $T$ denoise steps.}
\label{tab:accel_rho}
\small
\setlength{\tabcolsep}{4pt}
\begin{tabular*}{\columnwidth}{
    @{\extracolsep{\fill}}
    llrccc
}
\toprule
Model & Benchmark
& $n$ chunks
& $\rho_{\mathrm{full}}$
& $\rho_{\mathrm{best}}$
& $p^{\star}/T$ \\
\midrule

\multirow{3}{*}{$\pi_{0.5}$}
    & D3IL     & 712    & $0.826$ & $\mathbf{0.844}$ & 9/10 \\
    & LIBERO   & 32,647 & $0.541$ & $\mathbf{0.792}$ & 5/10 \\
    & RoboCasa & 8,277  & $0.381$ & $\mathbf{0.684}$ & 3/10 \\

\midrule
\multirow{3}{*}{SmolVLA}
    & D3IL     & 576    & $0.725$ & $\mathbf{0.834}$ & 7/10 \\
    & LIBERO   & 16,157 & $0.523$ & $\mathbf{0.638}$ & 6/10 \\
    & RoboCasa & 21,657 & $0.639$ & $\mathbf{0.816}$ & 4/10 \\

\midrule
\multirow{3}{*}{GR00T N1.7}
    & D3IL     & 732  & $0.622$ & $\mathbf{0.642}$ & 3/4 \\
    & LIBERO   & 23,273 & $0.656$ & $\mathbf{0.656}$ & 4/4 \\
    & RoboCasa & 14,613 & $0.565$ & $\mathbf{0.592}$ & 3/4 \\

\midrule
\multirow{2}{*}{VLA-JEPA}
    & LIBERO   & 26,980 & $0.679$ & $\mathbf{0.679}$ & 4/4 \\
    & RoboCasa & 13,728 & $0.547$ & $\mathbf{0.586}$ & 3/4 \\

\midrule
Toy & D3IL     & 673    & $0.680$ & $\mathbf{0.726}$ & 8/10 \\

\bottomrule
\end{tabular*}
\end{table}


\begin{table*}[!t]
\centering
\caption{
\textbf{Online failure detection across four FM models, two benchmarks and a real robot.}
Entries are \textbf{TPR$_{\pm\text{std}}$\,$/$\,median detection lead}: true positive rate at a target false-alarm rate $\alpha{=}0.1$, and the lead counting re-planning steps between the alarm and the end of the failing episode.
Mean and sample standard deviation are over $5$ random repeats ($10$ for the real robot). *Note that GR00T-N1.7 is severely undertrained on LIBERO-all.
The \textbf{Real robot} column is a $\pi_{0.5}$ policy on a real bimanual robot.}
\label{tab:failure_detection_cusum}
\small
\setlength{\tabcolsep}{3.0pt}
\renewcommand{\arraystretch}{1.12}
\centering
\resizebox{\textwidth}{!}{%
\begin{tabular}{lcccccccccc}
\toprule
&
\multicolumn{4}{c}{\textbf{LIBERO-all}}
&
\multicolumn{4}{c}{\textbf{RoboCasa Atomic-Seen}}
&
\multicolumn{1}{c}{\textbf{Real robot}}
&
\\
\cmidrule(lr){2-5}
\cmidrule(lr){6-9}
\cmidrule(lr){10-10}

\textbf{Detector}
&
\shortstack{$\pi_{0.5}$\\{\scriptsize Succ.\ 96.3\%}}
&
\shortstack{SmolVLA\\{\scriptsize Succ.\ 80.6\%}}
&
\shortstack{GR00T-N1.7\\{\scriptsize Succ.\ 68.1\%*}}
&
\shortstack{VLA-JEPA\\{\scriptsize Succ.\ 94.1\%}}
&
\shortstack{$\pi_{0.5}$\\{\scriptsize Succ.\ 77.1\%}}
&
\shortstack{SmolVLA\\{\scriptsize Succ.\ 65.6\%}}
&
\shortstack{GR00T-N1.7\\{\scriptsize Succ.\ 62.5\%}}
&
\shortstack{VLA-JEPA\\{\scriptsize Succ.\ 68.1\%}}
&
\shortstack{$\pi_{0.5}$\\{\scriptsize 54 succ / 48 fail}}
&
\shortstack{\textbf{Average}\\{\scriptsize 9 settings}}
\\
\midrule
\multicolumn{11}{l}{\emph{Geometry-based cost-free methods \textbf{(Ours)}}} \\
\textbf{Accel}
& $\underline{0.85}_{\pm.02}$\,/$19$
& $\mathbf{0.87}_{\pm.01}$\,/$18$
& $0.53_{\pm.02}$\,/$18$
& $0.82_{\pm.02}$\,/$28$
& $\underline{0.49}_{\pm.02}$\,/$21$
& $\underline{0.77}_{\pm.03}$\,/$34$
& $0.55_{\pm.03}$\,/$26$
& $0.40_{\pm.02}$\,/$23$
& $\bold{0.94}_{\pm.06}$
& $\mathbf{0.69}$
\\
\textbf{Straightness}
& $0.78_{\pm.02}$\,/$18$
& $\mathbf{0.87}_{\pm.02}$\,/$17$
& $0.58_{\pm.01}$\,/$18$
& $0.64_{\pm.01}$\,/$24$
& $0.43_{\pm.03}$\,/$17$
& $\mathbf{0.83}_{\pm.02}$\,/$32$
& $\underline{0.60}_{\pm.03}$\,/$26$
& $0.42_{\pm.03}$\,/$22$
& $0.72_{\pm.05}$
& $\underline{0.65}$
\\
\midrule
\multicolumn{11}{l}{\emph{Resampling-based methods}} \\
\textbf{ACE}
& $\underline{0.85}_{\pm.01}$\,/$19$
& $0.64_{\pm.02}$\,/$16$
& $\underline{0.68}_{\pm.01}$\,/$21$
& $0.66_{\pm.02}$\,/$29$
& $0.30_{\pm.02}$\,/$23$
& $0.25_{\pm.02}$\,/$41$
& $\mathbf{0.62}_{\pm.03}$\,/$25$
& $0.42_{\pm.04}$\,/$27$
& $0.79_{\pm.04}$
& $0.58$
\\
\textbf{STAC}
& $\mathbf{0.90}_{\pm.02}$\,/$22$
& $\underline{0.77}_{\pm.01}$\,/$19$
& $0.62_{\pm.02}$\,/$23$
& $\underline{0.85}_{\pm.01}$\,/$26$
& $0.06_{\pm.01}$\,/$27$
& $0.56_{\pm.03}$\,/$35$
& $0.54_{\pm.04}$\,/$23$
& $\underline{0.49}_{\pm.03}$\,/$19$
& $0.38_{\pm.08}$
& $0.57$
\\
\textbf{Diff-DAgger}
& $0.67_{\pm.04}$\,/$20$
& $0.65_{\pm.01}$\,/$17$
& $0.51_{\pm.01}$\,/$18$
& $\mathbf{0.86}_{\pm.01}$\,/$29$
& $0.31_{\pm.02}$\,/$25$
& $0.51_{\pm.03}$\,/$37$
& $0.50_{\pm.01}$\,/$28$
& $0.16_{\pm.02}$\,/$30$
& $0.77_{\pm.05}$
& $0.55$
\\
\midrule
\multicolumn{11}{l}{\emph{Training-based methods}} \\
\textbf{FIPER}
& $0.81_{\pm.03}$\,/$19$
& $0.59_{\pm.06}$\,/$13$
& $0.67_{\pm.01}$\,/$21$
& $0.73_{\pm.07}$\,/$29$
& $0.33_{\pm.01}$\,/$21$
& $0.25_{\pm.02}$\,/$42$
& $\underline{0.60}_{\pm.03}$\,/$26$
& $0.42_{\pm.04}$\,/$27$
& $0.77_{\pm.05}$
& $0.57$
\\
\textbf{RND-OE}
& $0.38_{\pm.11}$\,/$18$
& $0.31_{\pm.05}$\,/$26$
& $0.63_{\pm.04}$\,/$18$
& $0.53_{\pm.09}$\,/$21$
& $0.30_{\pm.06}$\,/$18$
& $0.44_{\pm.06}$\,/$33$
& $0.20_{\pm.01}$\,/$28$
& $0.39_{\pm.03}$\,/$21$
& ${0.85}_{\pm.03}$
& $0.45$
\\
\textbf{LogpZO}
& $0.38_{\pm.08}$\,/$20$
& $0.31_{\pm.06}$\,/$27$
& $0.65_{\pm.03}$\,/$17$
& $0.46_{\pm.10}$\,/$20$
& $0.36_{\pm.07}$\,/$20$
& $0.48_{\pm.07}$\,/$34$
& $0.21_{\pm.02}$\,/$30$
& $0.42_{\pm.05}$\,/$22$
& $\underline{0.89}_{\pm.02}$
& $0.46$
\\
\textbf{SAFE}
& $0.80_{\pm.09}$\,/$19$
& $0.65_{\pm.08}$\,/$17$
& $\mathbf{0.70}_{\pm.05}$\,/$23$
& $0.82_{\pm.06}$\,/$31$
& $\mathbf{0.83}_{\pm.07}$\,/$18$
& $\underline{0.75}_{\pm.07}$\,/$29$
& $0.23_{\pm.08}$\,/$26$
& $\mathbf{0.59}_{\pm.07}$\,/$27$
& $0.38_{\pm.19}$
& $0.64$
\\
\bottomrule
\end{tabular}
}
\end{table*}

\paragraph{Accel rank-tracks the resample divergence.}
Table~\ref{tab:accel_rho} shows the $\rho$ between the resample divergence and both the full trajectory $\mathrm{accel}_T$ and the best prefix $\mathrm{accel}_{p^\star}$.
Exactly as the local-monotonicity predicts, 
the rank correlation $\rho$ and $\rho_p$ are positive across all 12 model$\times$benchmark cells, 
indicating that $\mathrm{accel}$ encodes common information about FM uncertainty agnostic to model architecture and data distribution.
We also observe that the best prefix $\mathrm{accel}_{p^\star}$ usually carries higher correlation than full $\mathrm{accel}$, more detail about this phenomenon deferred to Appendix.

Remarkably, a cost-free geometric read from a single denoising path largely recovers the ranking produced by expensive resampling, and effectively generalizes to different FM models.
Figure~\ref{fig:accel_rho} intuitively demonstrates this correlation with a $\pi_{0.5}$$\times$D3IL run.
We color the end-effector trajectory by $\mathrm{accel}$ and plot the resampled chunks in light gray at every chunk re-plan.
At low $\mathrm{accel}$ chunks like $2$ and $3$ (light blue), the resampled branches stay bundled in a tight cluster, indicating near-deterministic action generation; 
at high $\mathrm{accel}$ chunks like $0$ and $5$ (dark blue), they fan out widely across the space, revealing decision points at which the policy must choose among multiple plausible behaviors, 
such as alternative routes through the obstacle field.

\section{A Free Lunch in Online Failure Detection}
\label{sec:4_failure}

\subsection{Problem Formulation and Baselines}
Online failure detection aims to detect when a robot policy fails during task execution.
At timestep $t$, a VLA receives an observation $o_t$ consisting of vision, a natural-language instruction and robot state.
It then predicts a chunk of actions for the next $H$ steps.
At the same time, an online failure detector emits a score $z_t$ indicating the likelihood of failure at this time. 
If the score $z_t$ exceeds a certain threshold $\eta$, a failure flag would be raised.

We consider two families of failure detection scores as baselines: resample-based consistency and training-based OOD detection.
\textbf{ACE}~\citep{roemer2025fiper} draws $K$ resampled chunks, bin the end effector positions on an \emph{fitted} adaptive grid and report entropy.
\textbf{STAC}~\citep{agia2024sentinel} computes the RBF-MMD between future window of the previous chunk and current $K$ chunk resamples. 
\textbf{Diff-DAgger}~\citep{lee2025diffdagger} draws $K$ $(t,\text{noise})$ pairs at generation, re-noise the action chunks, and average the CFM loss.
\textbf{logpZO}~\citep{xu2025faildetect} trains an FM model from observation embedding$\to$noise on success rollouts and performs OOD detection.
\textbf{RND-OE}~\citep{roemer2025fiper} performs random network distillation on the observation embedding. It trains a predictor to regress a frozen random target on success embeddings.
\textbf{FIPER}~\citep{roemer2025fiper}, combination of the RND-OE and ACE legs, requires \emph{fitting}.
\textbf{SAFE}~\citep{gu2025safe} trains a small MLP/LSTM probe on the hidden state of action expert, with labeled success and failure rollouts.

\subsection{Method}
\paragraph{Failure Score.}
All baselines above require additional data or computation, either through inference-time resampling or training on labeled data.
In contrast, we directly utilize $\mathrm{accel}$ as a free online failure score. At each action-chunk generation step, the policy produces a full $T$-step denoising trajectory, from which we compute $z_t=\mathrm{accel}$. The resulting stream ${z_t}_t$ serves as the online failure signal throughout the rollout.
We additionally introduce \textbf{straightness}, defined as the chord-to-arc ratio of the denoising path, as a free comparator capturing a similar geometric property of $\mathrm{accel}$.

 Both scores are computed only over the executed action window rather than the full action horizon, thereby aligning them with the actions actually applied. We use the second-to-last prefix $\mathrm{accel}_{-2}$ as truncated prefixes correlate stronger with posterior spread.

\paragraph{Calibrating threshold $\eta$.}
Most baseline including SAFE \citep{gu2025safe}, logpZO \citep{xu2025faildetect} and FIPER \citep{roemer2025fiper} use \emph{functional conformal prediction} (CP) to calibrate a time-indexed threshold $\eta_t$ for ${z_t}$. However we argue that such time-indexed CP degenerates on long-horizon rollouts, and instead we calibrate a time-invariant threshold $\eta$ on the peak of episode-level \emph{cumulative sum} (CUSUM) \citep{page1954continuous} $\max S_t$ from a calibration set $C$ of $M$ held-out successful rollouts. Detail discussed in Appendix.

Specifically, we apply a one-sided CUSUM procedure to ${z_t}$, and estimate a reference level $\mu_0$, score scale $\sigma_z$, and slack $k=c\,\sigma_z$ from $C$. The statistic evolves as
\begin{equation}
S_t=\max\bigl(0,\;S_{t-1}+z_t-\mu_0-k\bigr),\qquad S_0=0,
\label{eq:cusum}
\end{equation}
and an alarm is raised at the first timestep $t$ such that $S_t>\eta$. 

\begin{figure*}[!t]
    \centering
    \includegraphics[width=\textwidth]{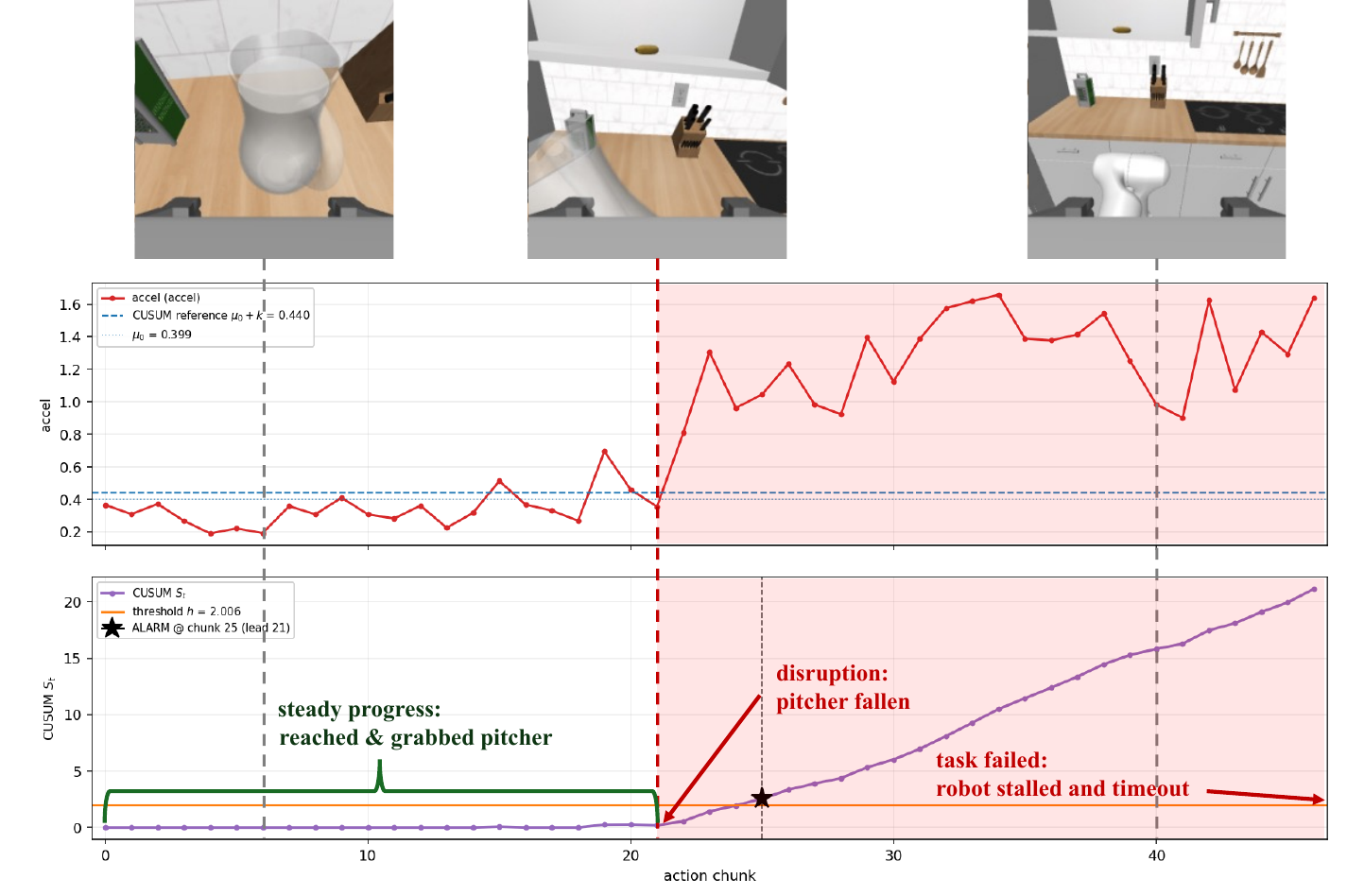}
    \caption{\textbf{$\mathrm{accel}$ and CUSUM time serties of a failed episode.} $\mathrm{accel}$ and CUSUM remain silent until disruption occurs and raise alarm within several chunks.}
    \label{fig:case_study_fail}
\end{figure*}

\subsection{Experiments}

\paragraph{Setup.} We experiment in both simulation and real robot execution. 

\noindent\textit{Simulation.} We use the eight model$\times$benchmark settings introduced in Section~\ref{sec:3_accel}. Each detector is evaluated on the same episodes and coupled with the same CUSUM monitor, using $c=0.25$ and conformal level $\alpha=0.1$. 
Within each run, we randomly draw a calibration set of $M{=}50$ held-out successful rollouts and yields its $\mu_0, k,\eta$.
We evaluate all baselines under a realistic deployment budget. Resample-based baselines use $K{=}32$ resamples per decision, while training-based baselines are fitted on $32$ successful rollouts or a balanced split of $16$ successful and $16$ failed rollouts for SAFE~\citep{gu2025safe}.

\noindent\textit{Real execution.} We evaluate on a real bimanual robot with a $\pi_{0.5}$ policy fine-tuned for medical-supply sorting. 
The calibration set is $M{=}8$ and the budgets are scaled proportionally: $K{=}16$, the fitted detectors use $5$ successes and SAFE a balanced $3{+}3$ split. Appendix~\ref{app:real-robot} gives the details.

The experiment is repeated $5$ times and we report the results in Table \ref{tab:failure_detection_cusum}.

\paragraph{Geometry-based scores are strong detectors.}
Across simulation and real-world execution, $\mathrm{accel}$ and Straightness match or outperform the baselines, ranking first and second in average TPR over the nine settings at a target FPR of $0.1$, ahead of the strongest baseline SAFE, which requires the most costly probe training on both successful and failed samples.
They raise alarms $17$ to $34$ action chunks before the failure timeout, $46\%$ to $65\%$ of max length.

Their advantage lies not only in detection performance but also in stability across both settings and random runs.
Across all model$\times$benchmark combinations, no baseline detector performs consistently well and every winning baseline degenerates on some specific settings. For example, STAC, a powerful baseline with 2 wins collapses to $0.06$ on $\pi_{0.5}\times$Atomic-Seen, because the policy executes its \emph{entire} predicted chunk before re-planning and leaves no future timestep which STAC depends on.
Meanwhile, $\mathrm{accel}$ and Straightness never severely collapse, suggesting high generalizability to different models and robustness to various failure modes with no specialized assumptions.

Furthermore, with the resampling or fitting budgets a real deployment can afford, most baselines suffer from a high deviation issue, especially training-based methods like SAFE. The free geometric scores are considerably more stable, with TPR standard deviations of only $0.01$--$0.03$. 

\paragraph{Where a geometric read is blind.}
The $\mathrm{accel}$ and Straightness are complementary measurements of the same geometric signature: how much the velocity varies along the denoising path. Both are most sensitive to failures preceded by posterior dispersion and may therefore miss cases in which their correlation with dispersion is weak or the policy commits confidently to an incorrect action generation.
Two mechanisms are at work: either there is nothing to see, the error lying outside the FM head; or the posterior is dispersed orthogonally to the realized path, which one trajectory cannot read (Appendix~\ref{app:faithfulness}).

Geometry-based methods degrade for undertrained models, as observed in GR00T-N1.7$\times$LIBERO-all. We hypothesize that insufficient training prevents the model from learning a stable FM field, resulting in noisy, irregular geometry that weakens the signal captured by $\mathrm{accel}$.
More broadly, almost all detectors degrade on RoboCasa, suggesting that the environment introduces more diverse and challenging failure modes that are intrinsically harder to identify. We leave the systematic investigation of the underlying causes of these degrades to future work.

\section{Case Studies}

\subsection{How $\mathrm{accel}$ Works as a Failure Score?}
We plot the time-series of $\mathbf{accel}$ and CUSUM of a failure and success rollout in Figure \ref{fig:case_study_fail} and \ref{fig:case_study_fail_2}.

During the failure rollout, the robot first reached for and grasped the pitcher, making steady task progress while both $\mathbf{accel}$ and CUSUM remained inactive. Around the 21st chunk, the pitcher slipped from the gripper, causing $\mathbf{accel}$ to rise sharply above its reference level and CUSUM crossed the threshold $\eta$ several chunks later. They clearly partitioned the rollout into normal and failure stages. 

In the successful rollout, CUSUM remained low throughout. Although $\mathbf{accel}$ occasionally exhibited transient spikes above the reference level, these were absorbed by the slack parameter $k$ and did not trigger an alarm.

\subsection{Where Is the Uncertainty? A Per-Arm Read}
\label{subsec:per_arm}

So far $\mathrm{accel}$ has been used as a scalar, it says that a chunk cannot be trusted, we further demonstrate that $\mathrm{accel}$ can reveal local uncertainty, i.e., \emph{which part} of it cannot be trusted. On our bimanual robot, the action space separates into two $8$-D blocks (seven joints and a gripper per arm). We take two per-arm reads, $\mathrm{accel}^{(\ell)}$ and $\mathrm{accel}^{(r)}$, each with its own CUSUM threshold. 

\begin{figure*}[!t]
    \centering
    \includegraphics[width=\textwidth]{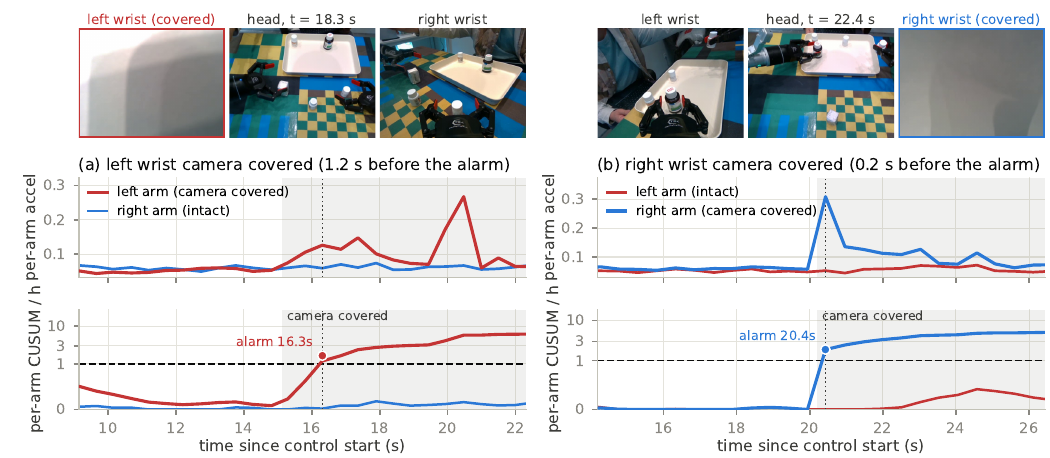}
    \caption{\textbf{Per-arm $\mathrm{accel}$ in wrist occlusion cases.}
    Two rollouts of the real-robot policy in which an operator covers one wrist camera mid-rollout.
    \textbf{(a)} Covering the left wrist camera: the covered arm largely stalls while the other keeps working, and $\mathrm{accel}^{(\ell)}$ departs from its success baseline while $\mathrm{accel}^{(r)}$ stays near its own.
    The left per-arm CUSUM crosses $\eta$ $1.2$~s after the cover, the right one only $16.5$~s later.
    \textbf{(b)} Covering the right wrist camera instead mirrors the attribution.}
    \label{fig:per_arm}
\end{figure*}

Wrist occlusion is a perfect demonstration of this decomposition: we randomly mask the wrist camera in the middle of execution and observe that the masked arm stalls while the other keeps progressing. As in  Figure~\ref{fig:per_arm}, once the left wrist camera is covered, $\mathrm{accel}^{(\ell)}$ climbs from $0.057$ to a mean of $0.095$ over the rest of the rollout, while $\mathrm{accel}^{(r)}$ bearly moves. The per-arm CUSUM inherits this asymmetry as the left arm crosses $\eta$ shortly after the cover, whereas the right arm alone would not fire. Covering the right wrist instead (Figure~\ref{fig:per_arm}b) mirrors the picture.

Removing one arm's close-up view widens the conditional distribution of \emph{that arm's} action coordinates while the other arm keeps its evidence, so the off-template excess is carried by the block whose covariance grew, and the other block stays close to its template.
The localization is therefore a direct read of the block structure of the FM posterior, and \textit{the cheapest one available}. This is the first instance of the finer-grained reading we believe the geometry supports, retaining not only \emph{whether} the denoising path bends but \emph{where} in the action space it bends.

\subsection{When $\mathrm{accel}$ Fails to Alarm?}
We further investigate the false negative (task failed w/o alarm) and false positive (task succeeded w/ alarm) episodes and identify the limitations of both $\mathbf{accel}$ and the environment's success criterion.

\paragraph{False negative: $\mathrm{accel}$-blind failure.}
Most FN cases fall into two categories:
\textbf{(1)} the first arises when the model \emph{confidently generates an incorrect action}, as discussed in the main text. In one case, the model misinterprets the instruction and turns on the wrong burner instead of turning off the correct one. $\mathbf{accel}$ remains low, suggesting that the policy is confident in this decision and that resampling from the FM head is unlikely to correct it. This error originates from high-level reasoning or instruction understanding in the VLM backbone, rather than from uncertainty in the FM head, and is therefore not reflected in FM geometry.
\textbf{(2)} the second arises in \emph{precision-sensitive manipulation}, such as pressing a small microwave button or placing a lid precisely on a blender. In these tasks, even minor action errors or jitter can cause failure. However, the resulting geometric deviations may be too subtle for $\mathbf{accel}$ to distinguish from the sea of background noise.

\paragraph{False positive: success-criterion-blind failure.}
We argue that many false positives reflect limitations of the simulator's success criterion rather than failures of $\mathbf{accel}$.
These cases often involve self-correction or temporary stalls. In complex tasks, the policy may require multiple attempts to grasp a fallen object or pause briefly to locate a target out of sight. We refer to such periods as \textbf{non-lethal failures}: the policy temporarily makes no task progress but eventually recovers.
Because the simulator evaluates only the final environment state, these transient failures are not captured by its success label. In contrast, $\mathbf{accel}$ detects these intermediate disruptions, providing a denser characterization of rollout dynamics.

\paragraph{Environment mislabeling.}
We also observe a small number of episodes in which the model completes the task but fails to trigger RoboCasa's success criterion. For some tasks, the criterion appears overly strict or misaligned with human judgment.

\section{Related Work}
\label{sec:5_related_work}

\paragraph{Flow Matching Model.}
Flow matching (FM) policies \citep{lipman2023flow, DBLP:conf/iclr/LiuG023} have become the default action head for embodied policies. 
They can generate precise and smooth action trajectories, capture complex multimodal action distributions
\citep{DBLP:journals/corr/abs-2410-24164, braun2024riemann},
and enable lower-latency sampling than diffusion policies
\citep{DBLP:conf/rss/ChiFDXCBS23}.
FM heads are adopted in multiple different models: $\pi$ family \citep{DBLP:journals/corr/abs-2410-24164, DBLP:journals/corr/abs-2504-16054} and GR00T family \citep{DBLP:journals/corr/abs-2503-14734} attaches a FM action head to a VLM in a mixture of experts style or a dual-system compositional architecture.
Recent works further couple FM head with world model components \citep{DBLP:journals/corr/abs-2504-02792, zhou2026tau0wm}, or reshape the denoising path for real-time execution \citep{jiang2025streaming}.

\paragraph{Uncertainty Quantification.}
Uncertainty quantification (UQ) gains much attention since generative AI tends to fail silently \citep{DBLP:journals/nature/FarquharKKG24, DBLP:journals/corr/abs-2406-15927, DBLP:conf/iclr/DaiXYLX25}. 
For FM models, multiple resample-based methods estimate uncertainty from the dispersion of Monte-Carlo resamples \citep{jazbec2025generative, DBLP:journals/corr/abs-2606-18043} or Bayesian inference \citep{kou2024bayesdiff}. 
Other work exploits the internal structure of FM models to estimate the uncertainty \citep{han2026uaflow}.
Closely related to us, FreeHunch \citep{DBLP:conf/iclr/RissanenHS25} and the concurrent work of \citet{DBLP:journals/corr/abs-2605-00941} share the intuition that uncertainty is intrinsically encoded in FM field geometry. Our work differs in both objective and realization, requiring no Jacobian estimation or additional probes, and is designed for FM-based embodied policies.

\paragraph{Online Failure Detection.}
Runtime failure detection for robot policies mainly fall into two categories.
\emph{Resample-based} methods like STAC~\citep{agia2024sentinel} and ACE~\citep{roemer2025fiper} measure the divergence of $K$ resampled action chunks; FIPER~\citep{roemer2025fiper} pairs that leg with an offline-fitted one and therefore pays both costs. Diff-DAgger~\citep{lee2025diffdagger} reads the self-reconstruction loss under re-noised chunks, a similar metric as divergence.
\emph{Training-based} detectors train an auxiliary model on labeled data and flag OOD embeddings, either with a learned density \citep{xu2025faildetect} or random network distillation \citep{roemer2025fiper} from success rollouts. 
Some supervised methods \citep{gu2025safe, park2026hideandseek} train small classifiers on the hidden state or action embedding to predict whether the rollout succeeded or not.
However, these detectors pays either an inference-time resampling or an offline training cost, while we provides a cost-free, real-time monitor that matches or exceeds these baselines.

\section{Conclusions}
\label{sec:6_conclusion}

We studied how uncertainty is expressed in the geometry of the flow field and grounded it to the deviation from an ideal template.
This connection led to $\mathrm{accel}$, a free proxy of uncertainty that reads the bending of an denoising path. As a downstream validation, we used it in failure detection and found that it provides stable and effective warning signals.

\paragraph{Limitations and Future Work.}
Our failure-detection study is not intended to pursue a new SOTA method but to prove the utility of $\mathrm{accel}$ as an uncertainty indicator.
It remains a rather coarse statistic for failure detection that misses confidently wrong actions and degrades with under-fit models. 
Future work can develop finer-grained reads from the geometry perspective by retaining when, where, and which dimensions the trajectory bends. We hope this work provides a simple geometric foundation to build 
more precise and capable uncertainty monitors.

\appendix
\section{Time-Indexed Conformal Band Degenerates on Long-Horizon Tasks}
\label{app:cp}

This appendix expands the argument of Section~\ref{sec:4_failure} for conformalizing a scalar CUSUM height instead of a time-indexed band.

\subsection{Late Horizon Issue of a Time-Indexed Band}
Functional CP calibrates a time-indexed band of the failure score on a set of \emph{successful} rollouts.
It forms a mean profile $m_t$ and a per-step spread $\phi_t$ across the calibration episodes, then sets an upper band $b_t=m_t+q_{1-\alpha}\,\phi_t$, where $q_{1-\alpha}$ is the split-conformal quantile of the normalized calibration residuals.
At test time an episode is flagged the first time $z_t>b_t$. Because $m_t,\phi_t,b_t$ are all functions of an \emph{absolute} timestep $t$, \textbf{the calibration episodes must be aligned to a common clock despite their varying length}.
In practice this is done by right-padding every episode with its own terminal value up to the longest sequence in the evaluation set, a length set by the timed-out failures.

This construction is harmless at short horizons but expensive at long ones, because the two classes have systematically different lengths. For embodied manipulation in either simulator or real world, a successful rollout terminates as soon as the task is solved, whereas a failing rollout stalls and keeps running until timeout.
Let $L^{\star}$ be the length of the longest successful calibration rollout and $L_{\max}>L^{\star}$ the timeout.
The padded length is set by the timed-out failures, so the band is defined on $[0,L_{\max}]$, yet on the interval $(L^{\star},L_{\max}]$ \emph{no successful rollout contributes a single real observation}, and every calibration entry there is a frozen copy of a terminal value.

This issue leads to three consequences. First, for $t>L^{\star}$ both $m_t$ and $\phi_t$ are computed entirely from held-last constants, so the band collapses to a flat horizontal line that reflects none of the long-horizon dispersion. Second, the empirical coverage on that interval has no genuine support. Third, and most fundamentally, calibration and test are no longer exchangeable there, the calibration sample consists of synthetic constants while the test sample is a genuinely running episode, so the two are not draws from a common distribution.
Therefore, calibration of a time-indexed band is least trustworthy precisely in the late-horizon regime, exactly where long-horizon failures manifest and a detector is needed most.

\paragraph{Experiment evidence.}
The length asymmetry is a property of most benchmarks.
On our $\pi_{0.5}\times$LIBERO cell, successful rollouts last $15.5$ chunks on average, whereas the all failing rollouts run to the step limit of the environment: $28$, $30$ or $52$ chunks depending on the suite.
Drawing $M{=}50$ calibration successes yields a longest calibration success of $L^{\star}\approx38$--$39$ chunks on average, against the $52$-chunk timeout of the long-horizon suite.
Averaged over the calibration draws actually used in our experiments, $53\%$ of the failing rollouts then outlive $L^{\star}$ and $16\%$ of all failing decision steps fall in the unsupported region.

The effect is present in every one of the eight cells of Table \ref{tab:failure_detection_cusum}: between $26\%$ and $55\%$ of failing episodes outlive $L^{\star}$, and between $4\%$ and $19\%$ of failing decision steps land where the band would be pure padding.
It is carried by the long-horizon tasks: on the short LIBERO suites, whose $28$- and $30$-chunk limits usually fall below the pooled $L^{\star}$, the unsupported interval is empty for most calibration draws. That is exactly the point, the cost is invisible at short horizons and grows with the horizon, which is the regime these policies are moving towards.

\subsection{CUSUM Avoids Late-Horizon Issue}
The calibration of Section \ref{sec:4_failure} conformalizes the episode-level CUSUM peak $P^{(i)}=\max_t S^{(i)}_t$, a single scalar summary of an episode of any length. No alignment, padding, truncation or common-clock assumption is introduced anywhere, so the exchangeability the split-conformal bound needs is a statement about episodes rather than about timesteps, which is the assumption a deployment already makes. The accumulation additionally works with, rather than against the length asymmetry. A mild but persistent elevation over a long failing rollout keeps integrating until it crosses $\eta$, which is the regime in which a flat late-horizon band is least informative.

\paragraph{Calibrating the threshold $\eta$.}
We change the object CP calibrates to the episode peak of CUSUM.
For each calibration episode $i$ we reduce its whole stream to a \emph{single scalar}, the episode CUSUM peak $P^{(i)}=\max_t S^{(i)}_t$, and set $\eta$ to the $\lceil (M+1)(1-\alpha)\rceil$-th smallest of $\{P^{(1)},\dots,P^{(M)}\}$, i.e.\ the $46$-th of $50$ at our $M{=}50$, $\alpha{=}0.1$.
No common clock assumption is introduced, so the calibration above is defined on raw streams of any length and no padding. 
\section{Prefix $\mathrm{accel}$ Along the Denoising Path}
\label{app:prefix-accel-rho}

\begin{figure}
    \centering
    \includegraphics[width=\linewidth]{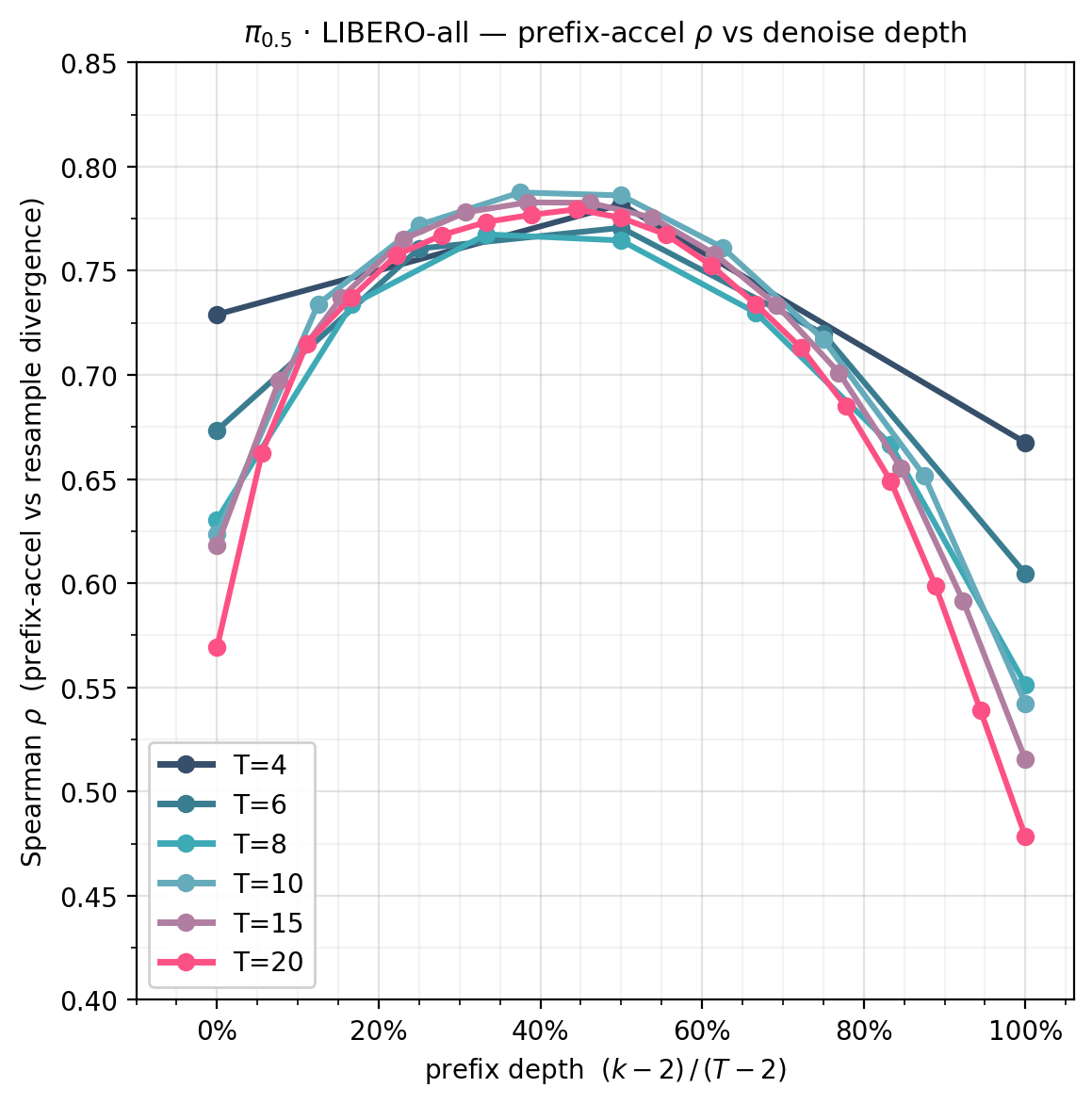}
    \caption{Prefix $\mathrm{accel}$'s $\rho$ vs prefix depth on $\pi_{0.5}$.}
    \label{fig:prefix_rho_steps}
\end{figure}

\begin{figure*}[!t]
    \centering
    \includegraphics[width=\textwidth]{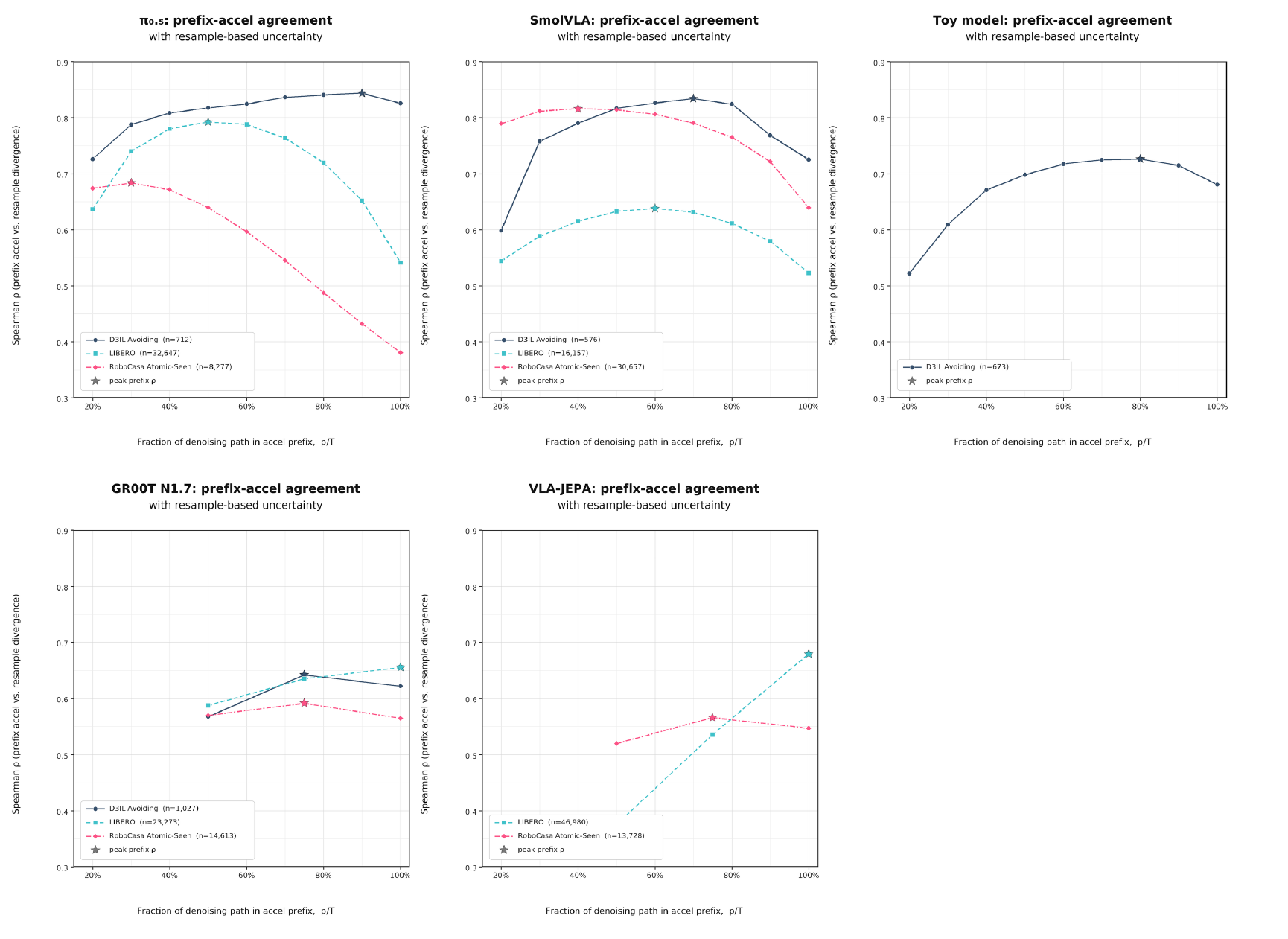}
    \caption{Prefix $\mathrm{accel}$'s $\rho$ of different models.}
    \label{fig:prefix_rho}
\end{figure*}

We complement the best-prefix results in Table \ref{tab:accel_rho} by examining how the agreement with resample-based uncertainty evolves over the denoising trajectory.  For each model--benchmark cell, we retain the same pooled set of chunks and the same resample-divergence reference, but recompute the proxy using only the first $p$ of the $T$ Euler steps.  We then report
\begin{equation}
    \rho_p=\operatorname{Spearman}\!\left(\mathrm{accel}_p,\,D_{\mathrm{resample}}\right),
    \qquad p=2,\ldots,T,
\end{equation}
where $p/T$ is the fraction of the noise-to-action denoising path included in the prefix.  
We plot each model separately in Figure \ref{fig:prefix_rho} and observe a common phenomenon across almost all models and benchmarks: the prefix curve rises with $p$, peaks in the middle of the denoise schedule, and then decays. 

\paragraph{Why a prefix $\mathrm{accel}$ wins: the terminal singularity.}

A striking regularity in Table~\ref{tab:accel_rho} is that the best correlation is attained strictly before the last step in $10$ of the $12$ cells.
The prefix curve rises with $p$, peaks in the middle of the denoise schedule, and then decays.
The decaying limb is exactly the terminal singularity analysed in Section~\ref{app:terminal}: the covariance term of Eq.~\eqref{eq:accel-cov} carries the schedule weight $s/(1-s)^{3}$, so as $s\!\to\!1$ (equivalently $t\!\to\!0$, the clean-action end) the bend is inflated by a diverging factor, and by Eq.~\eqref{eq:terminal-closed} the read saturates to an $O(1)$ value once the posterior spread becomes comparable to the schedule's terminal resolution.

Two further effects push the same way near $t=0$. The linear-interpolant CFM becomes singular there: when the data lie close to a low-dimensional manifold, the score $\nabla\log p_t$ grows rapidly and the velocity field becomes highly sensitive to small changes in its input. As a result, with a fixed-step Euler solver, the velocity differences in the last few steps contain not only useful posterior information, but also large discretization errors and numerical fluctuations. These changes can have large magnitudes without providing much additional information about $\Sigma$.

Therefore, including the final steps in $\mathrm{accel}_T$ can weaken the useful ranking information collected earlier. This is consistent with Table \ref{tab:accel_rho}. The final step can be an order of magnitude larger than the middle steps, while its correlation with $\Sigma$ is much weaker. The best correlation therefore appears before the endpoint.
In practice, we avoid this issue by using a prefix score $\mathrm{accel}_p$; the singularity itself is a property of the schedule and is avoided rather than removed.

The rising limb has a different origin, and it is worth separating the two. In the \emph{exact} field the ranking information is already present at $p=2$, so the rise cannot be an exact-field effect; it reflects the estimator's signal-to-noise ratio, since a short prefix sums only one or two velocity differences and a learned field contributes approximation error at every step, whereas the $\sigma$-dependent signal is concentrated late by the $(1-s)^{-3}$ weight. Accumulating more steps therefore improves the ratio until the terminal saturation takes over.

This decomposition also explains a signal we had left unaccounted for in Figure \ref{fig:prefix_rho}: the prefix correlation curves tend to peak earlier on the harder benchmark (RoboCasa) and later on easy tasks (D3IL), most visibly for the ten-step models $\pi_{0.5}$ and SmolVLA. A harder cell has a wider spread of posterior scales, so by the validity condition $\sigma\ll1-\tau$ of Proposition~A.4 its saturation reaches back to shallower prefixes, moving the optimum earlier. Task difficulty is in this sense encoded in the FM field and read off by $\mathrm{accel}_p$.

\paragraph{$\mathrm{accel}$ is robust against denoising resolution.} We rerun $\pi_{0.5}\times$LIBERO with different total denoising steps $T$ ($4-20$) and plot the prefix $\rho_p$ curves in Figure \ref{fig:prefix_rho_steps}. the x-axis is the relative prefix depth into the entire denoising path. Across all $T$, the curve demonstrates a common pattern to peak at around $40\%$ of the whole trajectory, reaching a similar best $\rho_{best}$ around $76\%-79\%$. This result indicates that $\mathrm{accel}$ is relatively robust to the denoising resolution, consistently carrying information about posterior divergence especially at the best step.
\section{A Complete Proof of $\mathrm{accel}$ Faithfulness}
\label{app:faithfulness}

\subsection{Complements to the Field-Level Theorems}
\label{app:field-complements}

We first record three statements deferred from Section~\ref{sec:2_geometry}, all immediate from the boxed identity \eqref{eq:tweedie-jacobian}.

\paragraph{The converse of Theorem 1.}
Suppose the excess $J+\tfrac{I}{1-s}$ vanishes for all $x$ at a single interior $s$. Then $\operatorname{Cov}(x_1\mid x_s=x)=0$ for every $x$, i.e.\ each posterior conditional is a point mass. That conditional is proportional to $p(x_1\mid o)\exp\bigl(-\lVert x-sx_1\rVert^2/2(1-s)^2\bigr)$, whose Gaussian factor is strictly positive and finite, so the conditional law is equivalent to $p(x_1\mid o)$ and carries the \emph{same support}. A point mass with that support forces the support to be a single point, hence $\Sigma=0$. So the affine-isotropic template is attained \emph{only} at maximal certainty, and contrapositively a nonzero excess at even one state implies $\Sigma\neq0$ --- which is the step Theorem~2 needs, and the direction a detector uses when it reads a small deviation as certainty. Note the quantifier: the hypothesis is over all $x$ at one $s$, not over one $x$.

\paragraph{The exact excess is symmetric PSD.}
Since $\operatorname{Cov}(x_1\mid x_s=x)\succeq0$ and is symmetric, \eqref{eq:tweedie-jacobian} forces $J+\tfrac{I}{1-s}\succeq0$ and symmetric for an \emph{exact} CFM field. A learned $v_\theta$ carries no such guarantee, so a skew-symmetric part or a negative eigenvalue in its excess cannot be read as an endpoint covariance at all: it is approximation, finite-data or schedule error. This is a free diagnostic, and it also bounds what the theory can claim about a learned model.

\paragraph{\eqref{eq:global-bound} is an exact decomposition.}
The same two lines that give the bound give an equality. Since $\mathbb{E}_{x_s}\mathbb{E}[x_1\mid x_s]=a^\star$, the left side of \eqref{eq:global-bound} equals $(1-s)^{-2}\operatorname{tr}\operatorname{Cov}(\mathbb{E}[x_1\mid x_s])$, and the law of total covariance $\Sigma=\mathbb{E}[\operatorname{Cov}(x_1\mid x_s)]+\operatorname{Cov}(\mathbb{E}[x_1\mid x_s])$ then yields
\begin{equation}
\operatorname{tr}\Sigma
=
\underbrace{\mathbb{E}_{x_s}\operatorname{tr}\operatorname{Cov}(x_1\mid x_s)}_{\text{still unresolved at time }s}
\;+\;
\underbrace{(1-s)^2\,\mathbb{E}_{x_s}\bigl[\lVert v-v_{\mathrm{aff}}\rVert^2\bigr]}_{\text{already legible in the field}} .
\end{equation}
The uncertainty thus splits exactly into what the field at time $s$ already reveals and what remains hidden in the conditional; the inequality in the main text keeps only the direction a detector needs.

\paragraph{What $\Sigma$ is, and is not.}
$\Sigma=\operatorname{Cov}(x_1\mid o)$ is the dispersion of the action distribution given the observation. It is therefore aleatoric in nature --- behavioural ambiguity and multimodality --- and carries no information about uncertainty over model parameters or model knowledge; upgrading \eqref{eq:tweedie-jacobian} to a statement about \emph{epistemic} uncertainty would require model stochasticity or a Bayesian model posterior, which we do not introduce. The OOD panel of Figure~\ref{fig:toy_flowfield} is accordingly an empirical observation: a learned field also departs from the template where it extrapolates, and by the previous paragraph that departure need not even be a valid covariance. Useful for monitoring, but a separate claim from Theorems~1--2.

\paragraph{Remark: $\mathrm{accel}$ is schedule-referenced, not an intrinsic curvature.}
We describe $\mathrm{accel}$ informally as reading how the path ``bends'', but $v'=x''$ is the \emph{parametrized} acceleration: geometric curvature retains only the component normal to the path, whereas $\mathrm{accel}$ also counts tangential speed changes. Under a reparametrization $s=\phi(r)$ of the denoise schedule the velocity becomes $\tilde v(r)=\phi'(r)\,v(\phi(r))$ and
\begin{equation}
\tilde v'(r)=\phi''(r)\,v(\phi(r))+\phi'(r)^2\,v'(\phi(r)),
\end{equation}
so the same geometric curve yields a different $\mathrm{accel}$. In $D{=}1$ every trajectory lies on a line, yet $\mathrm{accel}>0$ whenever $\Sigma\neq0$ --- see \eqref{eq:exact-family}. The honest description is therefore a \emph{schedule-referenced, normalized total variation of the velocity}, which is precisely why it is free to read off a fixed-step solver, and a different object from the parametrization-invariant chord-to-arc straightness used as a comparator in Section~\ref{sec:4_failure}.

\subsection{Setup}

This appendix collects the proofs behind the proxy $\mathrm{accel}$.
The object we reason about is the discrete read \eqref{eq:accel-disc} evaluated at the policy's \emph{own} step count $T$, which is what a deployment computes; the continuous functional \eqref{eq:accel-cont} is its $T\!\to\!\infty$ limit at a prefix $\tau=p/T$.
We state our results on a prefix held strictly inside $(0,1)$ because that limit, while it exists for each fixed posterior spread, is \emph{not uniform} in the spread at the clean-action end $s\!\to\!1$: the two limits $\sigma\!\to\!0$ and $T\!\to\!\infty$ do not commute there. Section~\ref{app:terminal} exhibits the phenomenon in closed form and shows why every statement below is stated either on the discrete path or on a prefix bounded away from $s=1$.
Away from that endpoint the whole argument rests on a single exact identity, from which the zero baseline, positivity, and local monotonicity follow in turn.
Throughout, $\lVert\cdot\rVert$ is the norm in the per-dimension standardized action coordinates the implementation uses; Section~\ref{app:whitened} records what that changes (almost nothing).

\subsection{The Trajectory Identity}

The proxy measures how much the denoising path bends, so we first compute the path acceleration $v'(s):=\tfrac{d}{ds}v(x(s),s)$ along the generative ODE. Recall from \eqref{eq:vmean} that the field points from the current state toward the posterior mean,
\begin{equation}
v(x,s)=\frac{m(x,s)-x}{1-s},\qquad m(x,s):=\mathbb{E}[x_1\mid x_s=x],
\label{eq:vm-recall}
\end{equation}
and that the denoising path solves $x'(s)=v(x(s),s)$, so the velocity along the path is the field $v$ itself. Write $m_{\mathrm{path}}(s):=m\bigl(x(s),s\bigr)$ for the posterior mean \emph{evaluated along the trajectory}---the moving point the denoiser is currently aiming at---so that $m_{\mathrm{path}}'(s)=(\partial_s m)+(\partial_x m)\,v$.

\begin{quote}
\noindent\textbf{Proposition~A.1 (Trajectory identity).}\;
\emph{Along any generative denoising trajectory $x'(s)=v(x(s),s)$ of \eqref{eq:vm-recall}, the rate of change of the velocity is}
\begin{equation}
v'(s)\;=\;\frac{m_{\mathrm{path}}'(s)}{1-s},
\label{eq:traj-identity}
\end{equation}
\emph{for an arbitrary $C^1$ posterior-mean field $m(x,s)$ on $s\in(0,1)$.}
\end{quote}
\noindent\emph{Proof.} Differentiate $v(x(s),s)=(m_{\mathrm{path}}(s)-x(s))/(1-s)$ in $s$, using $x'=v$:
\begin{equation}
v'
=\frac{m_{\mathrm{path}}'-x'}{1-s}
+\frac{m_{\mathrm{path}}-x}{(1-s)^2}
=\frac{m_{\mathrm{path}}'-v}{1-s}+\frac{v}{1-s}
=\frac{m_{\mathrm{path}}'}{1-s},
\label{eq:traj-proof}
\end{equation}
where the middle step substitutes $(m_{\mathrm{path}}-x)/(1-s)=v$ from \eqref{eq:vm-recall} and $x'=v$. \hfill$\square$

\noindent The identity is deceptively strong, and it is also \emph{distribution-free}: the representation \eqref{eq:vm-recall} follows from the linear interpolant and the independence of $x_0$ and $x_1$ alone, with no Gaussianity, so Proposition~A.1 and the two propositions that follow hold for any noise law. Gaussianity of $x_0$ enters only where the Tweedie identity \eqref{eq:tweedie-mean} is used, i.e.\ in the covariance handshake and in Proposition~A.4. Written naively the velocity's rate of change has two pieces, $v'=\partial_s v+J\,v$ (the explicit $s$-dependence plus the Jacobian $J=\partial_x v$ acting on the velocity), and both are individually nonzero even at perfect certainty. \eqref{eq:traj-proof} shows they cancel to the single term $m_{\mathrm{path}}'/(1-s)$. Motion toward a \emph{fixed} target contributes no acceleration, the path bends if and only if the posterior mean it is chasing is itself moving.

\paragraph{Handshake with the covariance of Section~\ref{sec:2_geometry}.}
The motion of $m_{\mathrm{path}}$ is not an unrelated quantity---it is Section~\ref{sec:2_geometry}'s covariance, read along the path. Since $v=(m-x)/(1-s)$, the field Jacobian $J$ and the posterior-mean Jacobian $M:=\partial_x m$ satisfy $M=I+(1-s)J=(1-s)\bigl(J+\tfrac{I}{1-s}\bigr)$; the bracket is the off-template excess of \eqref{eq:tweedie-jacobian}, so
\begin{equation}
M=(1-s)\Bigl(J+\tfrac{I}{1-s}\Bigr)=\frac{s}{(1-s)^2}\,\operatorname{Cov}(x_1\mid x_s=x).
\label{eq:M-cov}
\end{equation}
Substituting $m_{\mathrm{path}}'=\partial_s m+M\,v$ and \eqref{eq:M-cov} into \eqref{eq:traj-identity},
\begin{equation}
v'(s)=\frac{\partial_s m}{1-s}\;+\;\frac{s}{(1-s)^3}\,\operatorname{Cov}\bigl(x_1\mid x_s=x(s)\bigr)\,v(s).
\end{equation}
The geometric part of the trajectory's acceleration is the \emph{same conditional covariance} that Section~\ref{sec:2_geometry} equated with the off-template excess, now contracted with the velocity along the states the path actually visits. Where Section~\ref{sec:2_geometry} measured the departure of the whole \emph{field} from the template, Proposition~A.1 integrates that departure \emph{along one realized path}. This is why a single free trajectory can proxy a quantity defined over the entire field.

\subsection{Zero Baseline and Positivity}

\begin{quote}
\noindent\textbf{Proposition~A.2 (Zero Baseline).}\;
\emph{If $\Sigma=0$, then $m_{\mathrm{path}}'\equiv0$ and, by \eqref{eq:traj-identity}, $v'(s)\equiv0$, so the functional \eqref{eq:accel-cont} is $0$ for every $\tau\in(0,1)$. On the discrete path the metric \eqref{eq:accel-disc} is \emph{exactly} $\mathrm{accel}_p=0$ for every prefix $p$ and every step count $T$, not merely to $O(1/T)$.}
\end{quote}
\noindent\emph{Proof.} $x_1=a^{\star}$ almost surely, so $m(x,s)=a^{\star}$ for all $(x,s)$, $m_{\mathrm{path}}(s)\equiv a^{\star}$, $m_{\mathrm{path}}'\equiv0$, and \eqref{eq:traj-identity} gives $v'\equiv0$: the velocity along the path is constant.
For the discrete claim, note that the exactness is not automatic---a fixed-step solver generally perturbs the path---so we verify it by recursion. With $h=1/T$, $s_t=th$, $v_t=(a^{\star}-x_t)/(1-th)$ and $x_{t+1}=x_t+h\,v_t$,
\begin{equation}
a^{\star}-x_{t+1}=(a^{\star}-x_t)\Bigl(1-\tfrac{h}{1-th}\Bigr)=(a^{\star}-x_t)\,\tfrac{1-(t+1)h}{1-th},
\end{equation}
hence $v_{t+1}=\tfrac{a^{\star}-x_{t+1}}{1-(t+1)h}=\tfrac{a^{\star}-x_t}{1-th}=v_t$. Every summand of the numerator of \eqref{eq:accel-disc} vanishes identically. \hfill$\square$

\noindent Both \eqref{eq:accel-cont} and \eqref{eq:accel-disc} are ratios, so they require a non-degenerate denominator. Here that is the event $x_0\neq a^{\star}$, which has probability one; more generally we read $\mathrm{accel}$ as $0$ on the null set of stationary paths, and the implementation adds a fixed $\varepsilon$ to the denominator, which is the same convention.

\noindent This is the exact discrete counterpart of Theorem~1's straight, zero-curvature trajectory \eqref{eq:straightline}: perfect certainty is a flat-line velocity, and $\mathrm{accel}$ reads it as a hard $0$ at any denoising resolution, including the $T{=}4$ schedules in our evaluation.

\begin{quote}
\noindent\textbf{Proposition~A.3 (Positivity betrays uncertainty).}\;
\emph{If the path bends at any interior $s_0\in(0,1)$, i.e.\ $v'(s_0)\neq0$---or, on the discrete path, if $v_{t+1}\neq v_t$ for some $t$ so that $\mathrm{accel}_p>0$---then the posterior is non-degenerate, $\Sigma\neq0$.}
\end{quote}
\noindent\emph{Proof.} By \eqref{eq:traj-identity}, $v'(s_0)\neq0$ with $s_0\in(0,1)$ forces $m_{\mathrm{path}}'(s_0)\neq0$, so $m$ is not the constant field $a^{\star}$; the discrete case likewise contradicts the recursion of Proposition~A.2. Either way, by the contrapositive of Proposition~A.2 (if $\Sigma=0$ then $m\equiv a^{\star}$ and the discrete velocity is constant), $\Sigma\neq0$. \hfill$\square$

\noindent Propositions~A.2 and~A.3 give the qualitative statement the detector rests on: a certain condition produces a flat-velocity path, and a bending path betrays an uncertain one. Both are stated at \emph{interior} $s$, because at the boundary $s{=}0$ the schedule factor in \eqref{eq:accel-cov} vanishes and the field matches the template regardless of $\Sigma$.

\paragraph{Proposition A.3 does not reverse pointwise.}
It is worth being precise about how far the implication runs, because the natural converse is false \emph{on a single path}. At the level of the whole \emph{field} the converse does hold, by the support argument of Section~\ref{app:field-complements}: if the template is matched at every $x$ for one interior $s$, then $\Sigma=0$. But one realized path cannot see the whole field. Take $D\ge2$, $a^{\star}=0$ and an endpoint uncertain in a single direction, $\Sigma=\sigma^2uu^{\top}$ with $\lVert u\rVert=1$, and start from an $x_0$ exactly orthogonal to $u$. A short computation gives $m(x,s)=\tfrac{s\sigma^2}{s^2\sigma^2+(1-s)^2}\langle x,u\rangle\,u$, which vanishes identically along a path started orthogonal to $u$; that path is therefore exactly the point-mass path $x(s)=(1-s)x_0$, traversed at constant velocity, and
\begin{equation}
\mathrm{accel}=0\quad\text{while}\quad \Sigma=\sigma^2uu^{\top}\neq0 .
\end{equation}
So $\mathrm{accel}=0$ does not imply certainty pointwise: a trajectory reads the covariance only through $\operatorname{Cov}(x_1\mid x_s)\,v$, its action along the direction the path happens to travel, and is blind to the orthogonal complement. This is not a pathology to be hidden, it is exactly the $\lVert\Sigma_0x_0\rVert$ factor in the slope \eqref{eq:kappa} below: the offending $x_0$ are those in $\ker\Sigma_0$, a Lebesgue-null set, which is why the quantitative statement survives \emph{in expectation over the noise draw} while the pointwise equivalence does not. This is the honest reason $\mathrm{accel}$ is a monitor rather than an estimator of $\Sigma$: a single path probes the covariance in one direction, and it is the fresh noise draw at every replan that keeps the read informative on average.

\subsection{The Clean-Action End Is Singular}
\label{app:terminal}

Before quantifying \emph{how much} the path bends we must fix the domain, because the total variation of the velocity over the \emph{whole} schedule does not behave like the uncertainty. This is not a technicality: it is the reason the theory is stated on prefixes and, we will argue, the reason the best empirical prefix is interior.

Take the exactly solvable isotropic family: $a^{\star}=0$, $x_1\sim\mathcal{N}(0,\sigma^2 I)$, $x_0\sim\mathcal{N}(0,I)$. Then $x_s\sim\mathcal{N}\bigl(0,D(s)I\bigr)$ with $D(s)=(1+\sigma^2)s^2-2s+1$, the exact trajectory from $x_0$ is $x(s)=\sqrt{D(s)}\,x_0$, and, using the identity $2D''D-(D')^2\equiv4\sigma^2$,
\begin{equation}
v(s)=\frac{D'(s)}{2\sqrt{D(s)}}\,x_0,
\qquad
v'(s)=\frac{\sigma^2}{D(s)^{3/2}}\,x_0 .
\label{eq:exact-family}
\end{equation}
Since $D(0)=1$, $D(1)=\sigma^2$ and $\int_0^1 D^{-3/2}ds=\sigma^{-1}+\sigma^{-2}$, the numerator of \eqref{eq:accel-cont} at $\tau=1$ is
\begin{equation}
\int_0^1\lVert v'(s)\rVert\,ds=(1+\sigma)\,\lVert x_0\rVert,
\label{eq:terminal-num}
\end{equation}
which \emph{does not vanish} as $\sigma\to0^{+}$. The reason is geometric: for any $\sigma>0$ the flow must contract an $O(1)$ initial spread down to $O(\sigma)$, and it slightly overshoots, so the velocity \emph{reverses}, running from $-x_0$ at $s=0$ to $+\sigma x_0$ at $s=1$; its total variation is therefore $(1+\sigma)\lVert x_0\rVert$, an $O(1)$ change compressed into a window of width $O(\sigma)$ at the clean end. The same turning point fixes the denominator, which is the total variation of $\sqrt{D}$ and not an endpoint difference: $D'$ vanishes at $s^\star=1/(1+\sigma^2)$ with $\sqrt{D(s^\star)}=\sigma/\sqrt{1+\sigma^2}$, so the path contracts to $\tfrac{\sigma}{\sqrt{1+\sigma^2}}\lVert x_0\rVert$ and then re-expands to $\sigma\lVert x_0\rVert$, giving
\begin{equation}
\int_0^1\lVert v(s)\rVert\,ds=\Bigl[\bigl(1-\sqrt{D(s^\star)}\bigr)+\bigl(\sqrt{D(1)}-\sqrt{D(s^\star)}\bigr)\Bigr]\lVert x_0\rVert
=\Bigl(1+\sigma-\tfrac{2\sigma}{\sqrt{1+\sigma^2}}\Bigr)\lVert x_0\rVert,
\end{equation}
and dividing gives the closed form
\begin{equation}
\mathrm{accel}_{\tau=1}
=\frac{1+\sigma}{\,1+\sigma-2\sigma/\sqrt{1+\sigma^2}\,}
=1+2\sigma+O(\sigma^2).
\label{eq:terminal-closed}
\end{equation}
So the full-path \emph{continuum} functional is discontinuous at $\sigma=0$: its zero-uncertainty limit is $1$, not $0$, and its response is linear rather than quadratic in the spread. Equivalently, the leading-order description \eqref{eq:leading-vprime} of $\lVert v'\rVert$ is not uniform on $[0,1]$: it is valid only while $1-s\gg\sigma$, and its integral $\kappa(\tau)\sigma^2$ ceases to be small exactly in the window $1-\tau\lesssim\sigma$, where the exact read reaches an $O(1)$ value comparable to \eqref{eq:terminal-closed}.

Two remarks put this in its place. First, the exact zero baseline and the finiteness of the discrete read are untouched. Proposition~A.2 is unconditional---at $\Sigma=0$ the telescoping recursion gives $v_{t+1}=v_t$ for every $T$ and every prefix, with no window condition---and separately, a fixed-step solver reads its last velocity at $s=1-1/T$, so it stays clear of the terminal window whenever $\sigma\ll1/T$. Second, the correct reading is that the schedule tail is an \emph{amplifier} rather than a source: it inflates $\lVert v'\rVert$ by $(1-s)^{-3}$ without adding information about $\Sigma$. For a population with realistic spread the last steps do sit inside that amplification window---which is exactly the mechanism Appendix~\ref{app:prefix-accel-rho} observes, and the reason the best empirical prefix is interior rather than full. We therefore state the quantitative result on a prefix and treat the last step as the boundary case it is.

\subsection{Local Monotonicity}

Zero-versus-positive is not enough: the \emph{amount} of bending must track the \emph{amount} of uncertainty. We establish this at leading order in the posterior spread, which is the regime a detector operates in. Scale the posterior as $x_1=a^{\star}+\sigma\,\xi$ with $\xi$ centered, $\operatorname{Cov}(\xi)=\Sigma_0$ fixed and enough moments, so that $\Sigma=\sigma^2\Sigma_0$, and let $\sigma\to0$.

\begin{quote}
\noindent\textbf{Proposition~A.4 (Local monotonicity).}\;
\emph{Fix a prefix $\tau\in(0,1)$. In the small-spread regime $\sigma\ll1-\tau$ the expected proxy leaves zero strictly monotonically and linearly in the posterior variance,}
\begin{equation}
\mathbb{E}_{x_0}\bigl[\mathrm{accel}_\tau\bigr]\;=\;\kappa(\tau)\,\sigma^2\;+\;O(\sigma^3),
\label{eq:leading-order}
\end{equation}
\emph{with the explicit slope}
\begin{equation}
\kappa(\tau)=\frac{1}{2}\Bigl(\frac{1}{(1-\tau)^2}-1\Bigr)\,
\mathbb{E}_{x_0}\!\left[\frac{\lVert\Sigma_0 x_0\rVert}{\lVert x_0-a^{\star}\rVert}\right],
\qquad 0<\kappa(\tau)<\infty
\ \text{ whenever }\Sigma_0\neq0,
\label{eq:kappa}
\end{equation}
\emph{the finiteness holding in any action dimension $D\ge2$. Consequently there is a $\sigma^{\star}(\tau,\Sigma_0)>0$ such that $\sigma\mapsto\mathbb{E}_{x_0}[\mathrm{accel}_\tau]$ is strictly increasing on $(0,\sigma^{\star})$.}
\emph{For symmetric $\xi$---in particular for a Gaussian or a symmetric bimodal endpoint posterior---the error term is $O(\sigma^4)$.}
\end{quote}
\noindent\emph{Proof.} Conditioned on $x_s=x$, the law of $\xi$ has density proportional to $p_\xi(\xi)\exp\bigl(\tfrac{\sigma s\langle\xi,\,x-s a^{\star}\rangle}{(1-s)^2}-\tfrac{\sigma^2s^2\lVert\xi\rVert^2}{2(1-s)^2}\bigr)$, so expanding the exponential to first order in $\sigma$ and using $\mathbb{E}\xi=0$, $\operatorname{Cov}\xi=\Sigma_0$ gives
\begin{equation}
m(x,s)-a^{\star}=\frac{s\,\sigma^2}{(1-s)^2}\,\Sigma_0\,(x-s\,a^{\star})+O(\sigma^3),
\qquad
M=\frac{s\,\sigma^2}{(1-s)^2}\,\Sigma_0+O(\sigma^3),
\label{eq:mean-expansion}
\end{equation}
consistently with \eqref{eq:M-cov}, and independently of whether $p_\xi$ is Gaussian or multimodal; the $O(\sigma^3)$ term carries the third cumulant of $\xi$ and vanishes when $\xi$ is symmetric. The posterior mean is thus frozen at $a^{\star}$ to $O(\sigma^2)$, so the path is the affine one to the same order: writing $\delta:=x-x_{\mathrm{aff}}$, we have $\delta'=-\delta/(1-s)+(m_{\mathrm{path}}-a^{\star})/(1-s)$ with $\delta(0)=0$, and variation of constants gives $\delta(s)=(1-s)\int_0^s(m_{\mathrm{path}}-a^{\star})(1-r)^{-2}dr=O(\sigma^2/(1-s))$, which is $O(\sigma^2)$ on any prefix bounded away from $s=1$. Hence $x(s)-s\,a^{\star}=(1-s)\,x_0+O(\sigma^2)$, and because $m-a^{\star}$ in \eqref{eq:mean-expansion} already carries a factor $\sigma^2$, substituting the affine path there costs only $O(\sigma^4)$ and makes the moving target travel along a \emph{fixed} direction,
\begin{equation}
m_{\mathrm{path}}(s)-a^{\star}=\frac{s\,\sigma^2}{1-s}\,\Sigma_0 x_0+O(\sigma^3),
\qquad
m_{\mathrm{path}}'(s)=\frac{\sigma^2}{(1-s)^2}\,\Sigma_0 x_0+O(\sigma^3),
\end{equation}
so that Proposition~A.1 yields the leading-order path acceleration
\begin{equation}
v'(s)=\frac{\sigma^2}{(1-s)^3}\,\Sigma_0 x_0+O(\sigma^3),
\qquad \sigma\ll1-s .
\label{eq:leading-vprime}
\end{equation}
Because the direction $\Sigma_0x_0$ does not depend on $s$, the numerator of \eqref{eq:accel-cont} integrates in closed form, while the denominator is the mean affine-template speed:
\begin{equation}
\int_0^\tau\lVert v'\rVert\,ds=\frac{\sigma^2\lVert\Sigma_0x_0\rVert}{2}\Bigl(\frac{1}{(1-\tau)^2}-1\Bigr)+O(\sigma^3),
\qquad
\frac{1}{\tau}\int_0^\tau\lVert v\rVert\,ds=\lVert x_0-a^{\star}\rVert+O(\sigma^2).
\end{equation}
These expansions are pointwise in $x_0$: the small parameter is $\sigma\lVert x_0\rVert/(1-s)$, so the error is $O\bigl(\sigma^3(1+\lVert x_0\rVert^3)/(1-\tau)^3\bigr)$ uniformly on $[0,\tau]$. Since $\mathrm{accel}_\tau$ is dominated by $C(1+\lVert x_0\rVert)/\lVert x_0-a^{\star}\rVert$, which is integrable in the action dimension $D\ge2$, and the excluded event $\sigma\lVert x_0\rVert\gtrsim1-\tau$ has Gaussian-small probability $e^{-\Omega(((1-\tau)/\sigma)^2)}$, the expectation over the noise draw may be taken term by term; doing so gives \eqref{eq:leading-order} with the slope \eqref{eq:kappa}, the $O(\sigma^3)$ constant growing like $(1-\tau)^{-3}$. The slope is strictly positive without any genericity caveat: for $\Sigma_0\neq0$ the kernel of $\Sigma_0$ is a proper subspace and hence Lebesgue-null, so $\lVert\Sigma_0x_0\rVert>0$ almost surely, and likewise $\lVert x_0-a^{\star}\rVert>0$ almost surely, so the expectation of the ratio is positive. \hfill$\square$

\noindent Thus $\mathrm{accel}$ leaves $0$ \emph{linearly} in the posterior variance with a strictly positive slope: on the small-spread branch that a trained policy occupies, larger uncertainty yields a strictly larger expected proxy, which is exactly the rank-faithfulness a detector consumes. Because \eqref{eq:mean-expansion} never used the shape of $p_\xi$, the same statement covers a bimodal endpoint posterior whose mode separation grows from $0$: to leading order $\mathrm{accel}$ responds to the endpoint \emph{second moment} and does not distinguish a Gaussian posterior from a bimodal one of the same covariance. This is the aleatoric face of the very same covariance object as in Section~\ref{sec:2_geometry}, so $\mathrm{accel}$ responds to multimodality and to underfit variance alike---and, symmetrically, it is not a modality detector.

\paragraph{The discrete slope and its harmless $T$-dependence.}
Repeating the expansion on the fixed-step path gives the same law for the deployed read, $\mathbb{E}[\mathrm{accel}_T]=\kappa_T\,\sigma^2+O(\sigma^3)$, now with a slope that grows with the denoising resolution, $\kappa_T=\Theta(T^2)$: a finer grid resolves more of the $(1-s)^{-3}$-weighted bend near the clean end. Numerically fitting the exact-field Euler simulation of \eqref{eq:exact-family} over $\sigma\in[10^{-4},10^{-2}]$ recovers a $\sigma$-exponent of $2.00$ and a ratio $\kappa_T/T^2$ rising from $0.61$ at $T{=}4$ to $0.645$ at $T{=}40$; this exceeds the value $\tfrac12$ that continuing \eqref{eq:kappa} to $\tau=1-1/T$ would suggest, because inside the terminal window the fixed-step path itself deviates from the exact one at the same $O(\sigma^2)$ order as the signal --- which is why we quote only the order $\Theta(T^2)$ and not a closed-form constant. This growth is invisible to everything we report, because a $T$-dependent positive \emph{multiplicative} constant cancels in both consumers of the score: the Spearman $\rho$ of Table~\ref{tab:accel_rho} is rank-based, and the CUSUM statistic of Section~\ref{sec:4_failure} is standardized by the calibration reference $\mu_0$ and scale $\sigma_z$. It is the reason $\mathrm{accel}$ is empirically robust to the denoise-step count (Appendix~\ref{app:prefix-accel-rho}) even though its raw scale is not.

\paragraph{Sensitivity versus validity: why the best prefix is interior.}
Equations \eqref{eq:kappa} and Section~\ref{app:terminal} pull in opposite directions as the prefix deepens. The schedule factor $\tfrac12\bigl((1-\tau)^{-2}-1\bigr)$ \emph{increases} in $\tau$, so a deeper prefix is more sensitive to $\sigma$; but the validity window $\sigma\ll1-\tau$ \emph{shrinks}. Inside that window $\kappa(\tau)\sigma^2$ diverges while the exact read stays bounded at an $O(1)$ value of the same order as \eqref{eq:terminal-closed}: the response saturates and eventually turns non-monotone in $\sigma$, compressing ranks. The competition predicts an interior optimal prefix, and predicts that it moves \emph{earlier} for populations with larger spread. Both are what we observe: $p^{\star}/T$ in Table~\ref{tab:accel_rho} is $9/10$ on the easy D3IL cell, $5/10$ on LIBERO and $3/10$ on the hard RoboCasa cell.

\subsection{Standardized Coordinates}
\label{app:whitened}

The implementation divides each flattened action dimension by a fixed run-pooled scale before any norm is taken, so the $\lVert\cdot\rVert$ appearing in \eqref{eq:accel-cont} and \eqref{eq:accel-disc} is the corresponding weighted norm, $\lVert u\rVert_W^2=u^\top W u$ with $W$ a fixed positive diagonal. This is a re-weighting of the \emph{read}, not of the field or of the path: the trajectory the policy integrates is unchanged, so it is enough to carry $W^{1/2}$ through the norms and no step of the analysis has to be redone. Propositions~A.1--A.3 are statements about whether $v'$ and $v_{t+1}-v_t$ vanish, and $W^{1/2}$ is invertible, so they hold verbatim. In Proposition~A.4 only the two norms change, and \eqref{eq:leading-vprime} still supplies a single fixed direction, so
\begin{equation}
\kappa_W(\tau)=\frac{1}{2}\Bigl(\frac{1}{(1-\tau)^2}-1\Bigr)\,
\mathbb{E}_{x_0}\!\left[\frac{\lVert\Sigma_0x_0\rVert_W}{\lVert x_0-a^{\star}\rVert_W}\right]\;>\;0,
\end{equation}
which is \eqref{eq:kappa} with both norms weighted; strict positivity is unaffected. One practical consequence: because the scale is estimated by pooling, it must be supplied to the online detector as a \emph{fixed external reference} rather than re-estimated from the single chunk at hand, which is what we do.

\paragraph{Honest limitation of the guarantee.}
Our theoretical result is local and holds in expectation: under an exact CFM field, on a prefix bounded away from the clean-action end, and for a small-spread family with a fixed covariance shape, the expected $\mathrm{accel}$ increases with the posterior variance to leading order, with the explicit slope \eqref{eq:kappa}. Pointwise monotonicity is not merely left unproved but is \emph{false}: the read from a single noise draw is a stochastic, directionally incomplete observation of a field-level quantity, and the construction after Proposition~A.3 exhibits an $x_0$ for which $\mathrm{accel}=0$ at $\Sigma\neq0$. Global monotonicity across arbitrary distributions also does not hold, and Section~\ref{app:terminal} shows the response saturates once the spread is comparable to the schedule's terminal resolution. Since \eqref{eq:kappa} factorizes into the posterior scale $\sigma^2$ and a geometry term set by $\Sigma_0$ and $a^\star$, ranking across \emph{different} observations further assumes the geometry term varies little relative to the scale. Finally, the reference we correlate against is a diameter of the learned sampler's own endpoint pushforward rather than $\operatorname{Cov}(x_1\mid o)$ itself, and the two agree in rank only under the conditions given in Section~\ref{sec:3_accel}; acceleration in learned models may additionally contain approximation and discretization effects. These are the gaps the empirical study in Section~\ref{sec:3_accel} is there to cover.

\section{Toy Model Implementation Details}
\label{app:toy-model}


This section gives the complete specification of the toy experiment used in Figure~\ref{fig:toy_flowfield}. The toy is a learned conditional flow-matching
(CFM) model rather than an analytic vector field.  A single velocity network
$v_\theta(x_s,s,o)$ is shared by all conditions, where the action state is
$x_s\in\mathbb{R}^{2}$, the continuous observation is
$o\in\mathbb{R}^{8}$, and $s\in[0,1]$ denotes denoising progress from noise to
action.  Thus, the unimodal, multimodal, and held-out panels are different
observation-conditioned slices of the same learned field.

\paragraph{Observations and conditional targets.}
We first place six training observations at equally spaced angles
$\theta_i=2\pi i/6$ on the unit circle.  To avoid encoding the condition as a
discrete task identifier, the two-dimensional circle is lifted into an eight-dimensional observation space:
\begin{equation}
    o_i = Q
    \begin{bmatrix}\cos\theta_i & \sin\theta_i\end{bmatrix}^{\!\top},
    \qquad Q^\top Q=I_2,
    \label{eq:toy-observations}
\end{equation}
where $Q\in\mathbb{R}^{8\times2}$ is obtained once from the QR decomposition
of a seeded Gaussian matrix.  Consequently, observation norms and pairwise
distances on the circle are preserved.  Three of the six conditions are
randomly assigned a unimodal target and the other three a bimodal target.  For
every training condition, the clean-action distribution is a Gaussian mixture
\begin{equation}
        p(x_1\mid o_i)=\sum_{k=1}^{K_i}w_{ik}\mathcal{N}\!\left(x_1;\mu_{ik},\sigma_{ik}^{2}I_2\right),
        K_i\in\{1,2\}.
    \label{eq:toy-gmm}
\end{equation}
For each condition, mode means are sampled in
$[-2.3,2.3]^2$ by rejection sampling, with a minimum pairwise separation of
$2.0$ for bimodal targets.  Standard deviations are sampled independently and
uniformly from $[0.06,0.13]$.  Mixture weights are generated as
$w_i=\operatorname{softmax}(0.3z_i)$ with $z_i\sim\mathcal{N}(0,I_{K_i})$;
the unimodal weight is therefore one. We draw 4,096 samples from each
target once and freeze these finite sets, training only resamples from them
with replacement, and never queries the GMMs for fresh training samples.

To construct the epistemic condition, we additionally place three held-out
observations at midpoints of gaps between the six training angles and apply the
same lift $Q$.  These conditions lie on the observation manifold but outside
the discrete training support; we refer to them as OOD conditions in the main
text.  Crucially, they have no associated action target and contribute no
training examples.  Their fields and endpoint distributions are produced
solely by the network's continuous-condition generalization.  With seed 0, the
unimodal conditions are $o_1,o_2,o_5$, the bimodal conditions are
$o_0,o_3,o_4$, and the three held-out conditions are denoted
$o_{\mathrm{ood},0},o_{\mathrm{ood},1},o_{\mathrm{ood},2}$.

\paragraph{Velocity-network architecture.}
The scalar $s$ is mapped to a 64-dimensional sinusoidal embedding.  Specifically,
for $j=0,\ldots,31$, its sine and cosine features use frequency
$\exp[-(\log 10^4)j/32]$ and argument $2\pi s$.  The observation is encoded by
an $8\!\rightarrow\!32\!\rightarrow\!32$ MLP with a SiLU activation after
each linear layer.  We concatenate the current action state, time embedding,
and observation embedding, and pass the resulting 98-dimensional vector
through four 256-wide fully connected hidden layers with SiLU activations,
followed by a linear two-dimensional velocity output.  The network contains
224,578 trainable parameters and uses neither normalization layers nor dropout.

\paragraph{CFM objective and optimization.}
Each minibatch row first samples one of the six training observations uniformly,
then samples $x_1$ from that condition's frozen set.  We draw
$x_0\sim\mathcal{N}(0,I_2)$ and
$s\sim\mathcal{U}[10^{-3},1-10^{-3}]$, and use the linear interpolant and its
conditional velocity target
\begin{equation}
    x_s=sx_1+(1-s)x_0,
    \qquad u_s=x_1-x_0.
\end{equation}
During training only, the nominal observation is perturbed as
$\widetilde{o}=o+0.01\epsilon$, $\epsilon\sim\mathcal{N}(0,I_8)$.  The loss is
\begin{equation}
    \mathcal{L}(\theta)=
    \mathbb{E}\!\left[
    \left\|v_\theta(x_s,s,\widetilde{o})-(x_1-x_0)\right\|_2^2
    \right].
    \label{eq:toy-cfm-loss}
\end{equation}
We optimize it for 40,000 steps using Adam with batch size 1,024, learning rate
$10^{-3}$, and zero weight decay; no learning-rate schedule or EMA is used.
Training uses float32, seed 0, and a CUDA device (with a CPU fallback).  The
training loss is logged every 500 steps, and generation fidelity is evaluated
every 5,000 steps using 2,000 generated and target samples per in-distribution
condition.

\paragraph{Sampling and visualization.}
At test time we sample $x_0\sim\mathcal{N}(0,I_2)$ and integrate
$\mathrm{d}x/\mathrm{d}s=v_\theta(x,s,o)$ from $s=0$ to $1$ with 40 explicit
Euler steps of size $1/40$.  For the figure, each condition uses the same batch
of 512 initial noise samples (seed $0+4242$), so differences between panels
cannot be attributed to different noise draws.  The velocity streamplot is
formed by averaging the sampled positions and velocities on a
$26\times26$ grid and discarding cells with fewer than five samples.  The
endpoint posterior spread displayed in the figure is the mean of the two
coordinate-wise endpoint standard deviations.

The three displayed conditions are selected reproducibly from the nine-condition
pool: the unimodal condition with the smallest endpoint spread ($o_1$), the
bimodal condition with the largest endpoint spread ($o_3$), and the held-out
condition with the largest median trajectory-curvature score ($o_{\mathrm{ood},0}$).  Their endpoint spreads are 0.0815, 0.7524, and 0.4421, respectively.  


\section{Failure-Detection Experimental Details}
\label{app:failure-detection-details}

This section provides the implementation and data-splitting details for the online
failure-detection results in Table~\ref{tab:failure_detection_cusum}.  The purpose is to
make the comparison reproducible rather than to introduce an additional evaluation.

\paragraph{Evaluation grid and labels.}
We evaluate four flow-matching policies: $\pi_{0.5}$, GR00T N1.7, VLA-JEPA, and
SmolVLA on LIBERO-all and RoboCasa Atomic-Seen, general result listed in Table \ref{tab:failure-detection-population}.
LIBERO-all pools the LIBERO-10, Goal, Object, and Spatial suites.  Atomic-Seen contains
18 single-skill RoboCasa tasks evaluated in held-out target kitchens.  A detector emits
one scalar whenever the policy replans an action chunk; hence all alarm and lead-time
quantities below are measured in re-planning steps (action chunks), not simulator
timesteps.  We retain only the part of a predicted chunk that is actually executed.  The
executed-window lengths on LIBERO/Atomic-Seen are respectively $10/16$ for $\pi_{0.5}$,
$8/8$ for GR00T N1.7, $7/8$ for VLA-JEPA, and $10/10$ for SmolVLA.

We use environment success as the episode-level ground-truth label.  For LIBERO, the
per-rollout termination label was checked against the success array saved by the
evaluation loop.  For RoboCasa, labels come directly from the simulator-client success
records; rollout identity and the number of recorded chunks were cross-checked when the
distributed shards were merged.  Thus, no trajectory-length or detector-score proxy is
used to assign success and failure labels.

\paragraph{Detector inputs and budgets.}
The cost-free geometric scores are
$\mathrm{accel}$, Straightness.  ACE and STAC use Monte-Carlo action
resamples, and the flow-matching loss (reported as Diff-DAgger in the main text) uses
re-noised model evaluations.  The observation-embedding OOD family comprises RND-OE and logpZO.  FIPER combines its RND-OE and ACE legs, while SAFE is a supervised probe on the action expert's last-layer hidden feature.

All resampling-based scores consume the complete set of $K{=}32$ candidates stored at
each decision, no policy inference is repeated during detector scoring.  Geometry and
resample scores are computed on the executed window.  For $\mathrm{accel}$, action dimensions are standardized with a fixed, run-pooled scale. 

Within each repeat, the unsupervised embedding-OOD family and FIPER are fitted on 32
randomly selected successful rollouts. SAFE uses an independently drawn, balanced set
of 16 successful and 16 failed rollouts. We exclude the union of these two fit sets from
the scored population for \emph{every} detector, including methods that do not require
fitting, so detector comparisons within a repeat remain paired. The five fit and learned
model seeds are $3200,\ldots,3204$.  Candidate-subset seeds $6400,\ldots,6404$ are stored
for provenance, although they do not change the result when all 32 saved candidates are
used.

\begin{table*}[t]
    \centering
    \caption{}
    \label{tab:failure-detection-population}
    \small
    \setlength{\tabcolsep}{4.5pt}
    \begin{tabular}{llrrrr}
        \toprule
        Model & Benchmark & Episodes & Success & Failure & Realized FPR \tabularnewline
        \midrule
        $\pi_{0.5}$ & LIBERO-all
            & $1936.0$ & $1878.0$ & $58.0$ & $0.097$ \\
        VLA-JEPA & LIBERO-all
            & $1936.2$ & $1833.2$ & $103.0$ & $0.097$ \\
        GR00T N1.7 & LIBERO-all
            & $737.6$ & $506.6$ & $231.0$ & $0.094$ \\
        SmolVLA & LIBERO-all
            & $737.0$ & $598.0$ & $139.0$ & $0.101$ \\
        \midrule
        $\pi_{0.5}$ & Atomic-Seen
            & $836.0$ & $646.0$ & $190.0$ & $0.097$ \\
        GR00T N1.7 & Atomic-Seen
            & $298.0$ & $179.0$ & $119.0$ & $0.096$ \\
        VLA-JEPA & Atomic-Seen
            & $298.4$ & $199.4$ & $99.0$ & $0.100$ \\
        SmolVLA & Atomic-Seen
            & $737.0$ & $478.0$ & $259.0$ & $0.098$ \\
        \bottomrule
    \end{tabular}
\end{table*}

\paragraph{CUSUM calibration and metrics.}
For each detector and calibration draw, we pool the chunk scores from $M{=}50$ held-out
successful episodes to estimate the reference mean $\mu_0$ and standard deviation
$\sigma_z$.  We use the one-sided statistic
\begin{equation}
    S_t=\max\{0,S_{t-1}+z_t-\mu_0-c\,\sigma_z\},
    \qquad S_0=0,\quad c=0.25.
\end{equation}
For every calibration episode $i$, we compute its peak $P^{(i)}=\max_t S_t$ and set the
alarm height to the split-conformal order statistic
\begin{equation}
    r=\left\lceil(M+1)(1-\alpha)\right\rceil,
    \qquad \eta=P_{(r)}.
\end{equation}
With $M=50$ and $\alpha=0.1$, this is the 46th smallest calibration peak.

An alarm is the first chunk with $S_t>\eta$.  The true-positive rate (TPR) is the fraction
of failed episodes that alarm.  For detected failures, lead is the number of remaining
chunks, $L=T_{\mathrm{ep}}-1-t_{\mathrm{alarm}}$; we first take the median across detected
failed episodes. Across all 112 cell, detector aggregates, the median realized FPR is $0.097$ with interquartile range $[0.095,0.100]$, confirming that the methods are compared near the intended operating
point. 

\section{Real-Robot Evaluation Details}
\label{app:real-robot}

This section documents the real-robot experiment.
We deploy a bimanual manipulator with two $7$-DoF arms and parallel-jaw grippers, observed by
three $480{\times}640$ RGB cameras mounted on the chest and on each wrist. The policy is
$\pi_{0.5}$~\citep{DBLP:journals/corr/abs-2504-16054} fine-tuned for a medical-supply sorting task
and executed with $10$ denoising steps per decision. Control is asynchronous, the controller executes the current chunk and requests a
new one every $5$ steps, so consecutive chunks overlap by $11$ steps.  Every accepted request is
one decision, and only the executed part of each chunk is scored, as in simulation.

All $102$ rollouts come from a single session, so lighting, operator, object layout and camera extrinsics are shared between the success and the failure class. The success class has $54$ rollouts and the failure class has $48$ rollouts in four families: \emph{human intervention}, where the operator reaches into the workspace and moves
the scene mid-rollout; \emph{tray removal}, where the target tray is pulled away;
\emph{abnormal position}, where the supplies are placed in configurations absent from the
demonstrations; and \emph{wrist occlusion}, where the wrist camera is covered. 

The first $3$ chunks of every rollout are dropped before fitting and scoring. The reason is measurable: in this session, chunks $0$ and $1$ exceed the success steady-state $99$th percentile of $\mathrm{accel}$ in $47\%$ and $89\%$ of \emph{successful} rollouts, chunk $2$ in
$5\%$, and chunks $3$ onwards in none. This start-up transient is a property of the policy leaving its initial state, not evidence of failure, and keeping it would let CUSUM accumulate credit on every
rollout alike.
\section{Ablation Study on Failure Detection}

\begin{figure*}[!t]
    \centering
    \includegraphics[width=\textwidth]{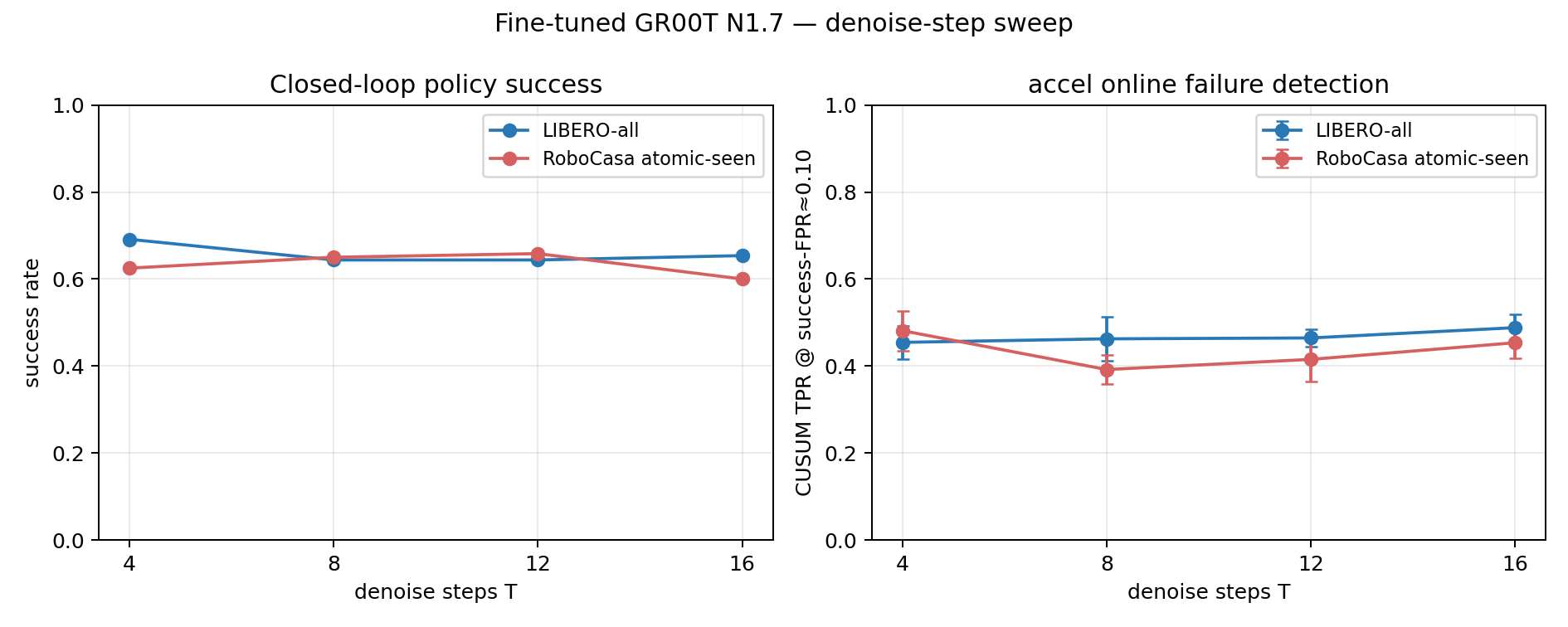}
    \caption{Prefix $\mathrm{accel}$'s $\rho$ of different models.}
    \label{fig:groot_step_sweep}
\end{figure*}

We sweep the total denoising step $T$ of GR00T-N1.7 and report the success rate and $\mathbf{accel}$ failure detection performance in Figure \ref{fig:groot_step_sweep}. No significant change observed in TPR across different $T$.

\begin{figure*}[!t]
    \centering
    \includegraphics[width=\textwidth]{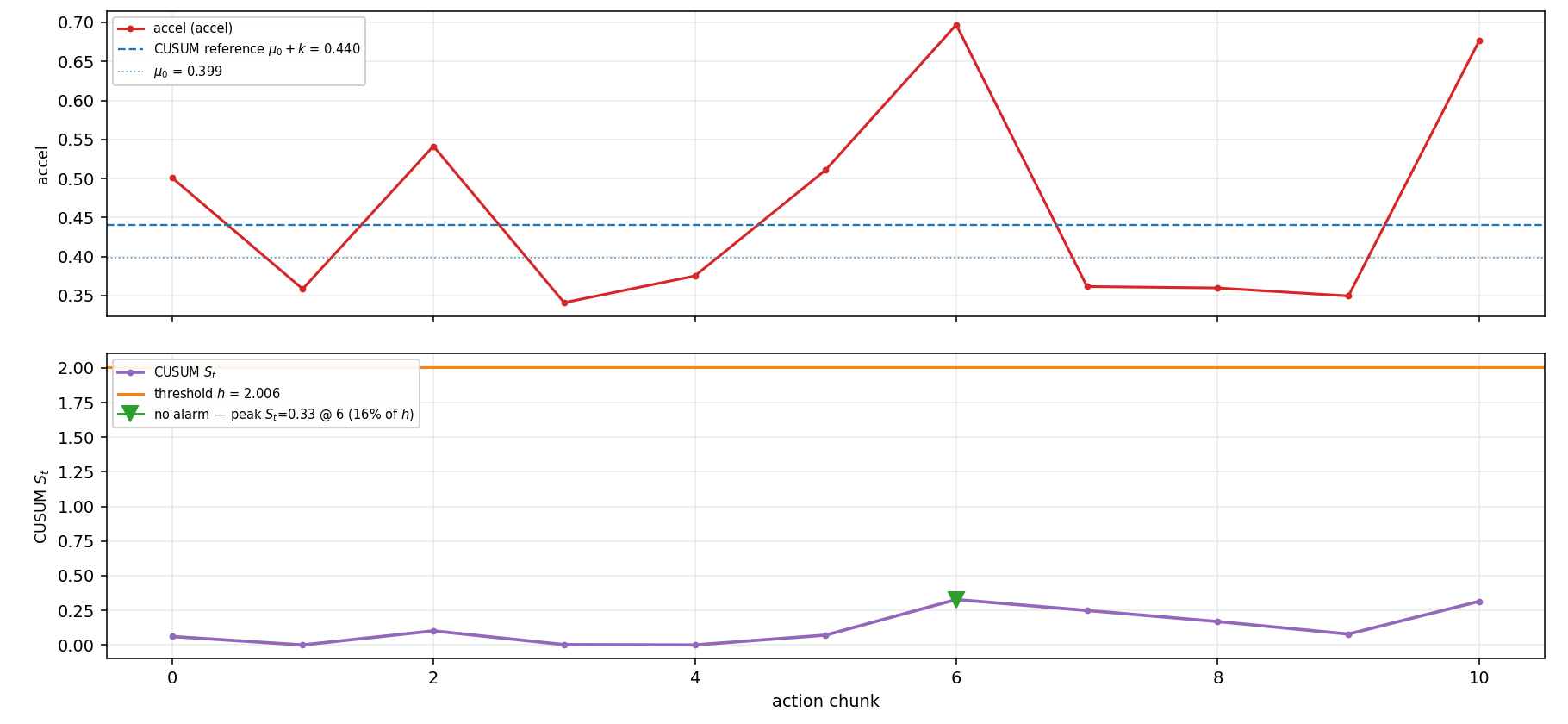}
    \caption{\textbf{$\mathrm{accel}$ and CUSUM time serties of a success episode.} $\mathrm{accel}$ and CUSUM remain silent throughout.}
    \label{fig:case_study_fail_2}
\end{figure*}

\bibliography{aaai2027}

\end{document}